\documentclass[sigconf]{acmart}

\AtBeginDocument{%
  \providecommand\BibTeX{{%
    \normalfont B\kern-0.5em{\scshape i\kern-0.25em b}\kern-0.8em\TeX}}}

\setcopyright{acmcopyright}
\copyrightyear{2020}
\acmYear{2020}
\acmDOI{10.1145/1122445.1122456}

\acmConference[Barcelona '20]{Barcelona '20: ACM Conference on Fairness, Accountability and Transparency}{Jan 27--30, 2020}{Barcelona, Spain}
\acmPrice{15.00}

\usepackage{subcaption}
\newcommand{\cP}{\mathcal{P}}
\newcommand{\cZ}{\mathcal{Z}}
\newcommand{\cK}{\mathcal{K}}
\newcommand{\cS}{\mathcal{S}}
\newcommand{\cI}{\mathcal{I}}
\newcommand{\bx}{\mathbf{x}}
\usepackage{listings}
\usepackage{lipsum}
\usepackage{courier}
\usepackage{afterpage}
\newcommand{\ignore}[1]{}

\newif\ifappendix
\appendixtrue
\renewcommand\footnotetextcopyrightpermission[1]{} % removes footnote with conference information in first column

\begin{document}

\title{One Explanation Does Not Fit All:\\
A Toolkit and Taxonomy of AI Explainability Techniques}
%[AD] Moving from Theory to Practice or Lab to Practice or (in) Lab to (on) 
%Field/Real World 
%[DW] One Explanation Does Not Fit All
%Taxonomies for Different Ways of Explaining and their Reduction to Practice in the AI Explainability 360 Toolkit}

%% "affiliation" and "affiliationmark" commands
\author{Vijay Arya, Rachel K. E. Bellamy, Pin-Yu Chen, Amit Dhurandhar, Michael Hind}
\author{Samuel C. Hoffman, Stephanie Houde, Q. Vera Liao, Ronny Luss, Aleksandra Mojsilovi\'c}
\author{Sami Mourad, Pablo Pedemonte, Ramya Raghavendra, John Richards, Prasanna Sattigeri}
\author{Karthikeyan Shanmugam, Moninder Singh, Kush R. Varshney, Dennis Wei, Yunfeng Zhang}
%\email{vijay.arya@in.ibm.com, rachel@us.ibm.com, pin-yu.chen@ibm.com, adhuran@us.ibm.com, hindm@us.ibm.com, shoffman@ibm.com} 
%\email{rluss@us.ibm.com}
%\email{aleksand@us.ibm.com}
%\email{sami.mourad@ibm.com}
%\email{ppedemon@ar.ibm.com}
%\email{rraghav@us.ibm.com}
%\email{ajtr@us.ibm.com} 
%\email{psattig@us.ibm.com} 
%\affiliation{IBM Research}
%\email{karthikeyan.shanmugam2@ibm.com} 
%\email{moninder@us.ibm.com}  
%\email{krvarshn@us.ibm.com}  
%\email{dwei@us.ibm.com}  
%\email{zhangyun@us.ibm.com}  

\affiliation{IBM Research}

\renewcommand{\shortauthors}{Arya et al.}

\begin{abstract}
As artificial intelligence and machine learning algorithms make further inroads into society, calls are increasing from multiple stakeholders for these algorithms to explain their outputs. At the same time, these stakeholders, whether they be affected citizens, government regulators, domain experts, or system developers, present different requirements for explanations. Toward addressing these needs, we introduce AI Explainability 360\footnote{The web demonstrations, tutorials, notebooks, guidance material as well as a link to the github repository (\url{https://github.com/IBM/AIX360/}) are available at \url{http://aix360.mybluemix.net/}.}, an open-source software toolkit featuring eight diverse and state-of-the-art explainability methods and two evaluation metrics. Equally important, we provide a taxonomy to help entities requiring explanations to navigate the space of explanation methods, not only those in the toolkit but also in the broader literature on explainability. For data scientists and other users of the toolkit, we have implemented an extensible software architecture that organizes methods according to their place in the AI modeling pipeline. We also discuss enhancements to bring research innovations closer to consumers of explanations, ranging from simplified, more accessible versions of algorithms, to tutorials and an interactive web demo to introduce AI explainability to different audiences and application domains. Together, our toolkit and taxonomy can help identify gaps where more explainability methods are needed and provide a platform to incorporate them as they are developed. 
\end{abstract}

%\begin{CCSXML}
%<ccs2012>
 %<concept>
  %<concept_id>10010520.10010553.10010562</concept_id>
  %<concept_desc>Computer systems organization~Embedded systems</concept_desc>
  %<concept_significance>500</concept_significance>
 %</concept>
 %<concept>
  %<concept_id>10010520.10010575.10010755</concept_id>
  %<concept_desc>Computer systems organization~Redundancy</concept_desc>
  %<concept_significance>300</concept_significance>
 %</concept>
 %<concept>
  %<concept_id>10010520.10010553.10010554</concept_id>
  %<concept_desc>Computer systems organization~Robotics</concept_desc>
  %<concept_significance>100</concept_significance>
 %</concept>
 %<concept>
  %<concept_id>10003033.10003083.10003095</concept_id>
  %<concept_desc>Networks~Network reliability</concept_desc>
  %<concept_significance>100</concept_significance>
 %</concept>
%</ccs2012>
%\end{CCSXML}

%\ccsdesc[500]{Computer systems organization~Embedded systems}



\maketitle

\pagestyle{plain} % removes running headers

\section{Introduction}
\label{sec:intro}
The increasing deployment of artificial intelligence (AI) systems in high
stakes domains has been coupled with an increase in societal demands
for these systems to provide explanations for their predictions.
This societal demand has already resulted in new regulations
requiring explanations \cite{gdpr-goodman,gdpr-wachter,SelbstP2017,illinois-2019}.
These explanations can allow users to gain insight into the system's
decision-making process, which is a key component in fostering trust
and confidence in AI systems~\cite{rsi,Varshney2019}.  

However, many machine learning techniques, which are responsible for
much of the advances in AI, are not easily explainable, even by
experts in the field.  This has led to a growing research community~\cite{KimVW2018},
with a long history, focusing on ``interpretable'' or ``explainable''
machine learning techniques.\footnote{We use the terms explainable and interpretable fairly interchangeably; some scholars make a strong distinction \cite{Rudin2019}.}  However, despite the growing volume of
publications, there remains a gap between what society needs and what
the research community is producing.

One reason for this gap is a lack of a precise definition of an
explanation. This has led to some ambiguity in
regulations~\cite{gdpr-goodman,gdpr-wachter,SelbstP2017,illinois-2019} and
researchers defining their own problem statements, as well as solutions. A reason for the lack of precise definition is that different people in different settings may require different kinds of explanations. %Given this need, we in this paper provide a taxonomy and an open source toolkit that considers the point of view of the many possible explanation consumers.
%We see this paper as a first step to help bridge this gap
%by providing common terminology and open source toolkit
%that considers the point of view of the many possible explanation
%consumers. 
%
%When interacting with machine learning predictions, consumers require
%different kinds of explanations depending on their persona
We refer to the people interacting with an AI system as \emph{consumers}, and to their different types as \emph{personas}
\cite{Hind2019}. For example, a doctor trying to understand an AI
diagnosis of a patient may benefit from seeing known similar cases
with the same diagnosis; a denied loan applicant will want to
understand the main reasons for their rejection and what can be done to
reverse the decision; a regulator, on the other hand, will want to
understand the behavior of the system as a whole to ensure that it
complies with the law; and a developer may want to understand where
the model is more or less confident as a means of improving its
performance.

Since there is no single approach to explainable AI that always
works best, we require organizing principles for the space of
possibilities and tools that bridge the gap from research to practice.
In this paper, we provide a taxonomy and describe an open-source toolkit to address the
overarching need, taking into account the points of view of the many possible explanation consumers. Our contributions are as follows:

\begin{table*}[t]
  \centering
  \small
  \begin{tabular}{|c|c|c|c|c|c|c|}
   \hline
 Toolkit & Data & Directly & Local & Global & Persona-Specific & Metrics\\
 & Explanations & Interpretable & Post-Hoc & Post-Hoc & Explanations & \\\hline
 AIX360 & \checkmark & \checkmark & \checkmark & \checkmark & \checkmark & \checkmark \\\hline
 Alibi~\cite{alibi} & &  & \checkmark &   &   &   \\\hline
Skater~\cite{skater} & & \checkmark & \checkmark & \checkmark & & \\\hline
H2O~\cite{h2o} & & \checkmark & \checkmark & \checkmark & & \\\hline
InterpretML~\cite{interpret} & & \checkmark & \checkmark &  \checkmark  &  & \\\hline
EthicalML-XAI~\cite{ethicalml} & & & & \checkmark & & \\\hline
DALEX~\cite{dalex} & & & \checkmark & \checkmark & & \\\hline
tf-explain~\cite{tfexplain} & & & \checkmark & \checkmark & & \\\hline
iNNvestigate~\cite{innvestigate} & & & \checkmark & & & \\\hline
  \end{tabular}
  \caption{Comparison of AI explainability toolkits.}
  \label{tab:compare}
\end{table*}

\begin{itemize}
    \item \emph{Taxonomy Conception:} we propose a simple yet comprehensive taxonomy of AI explainability that considers varied perspectives. This taxonomy is actionable in that it aids users in choosing an approach for a given application and may also reveal gaps in available explainability techniques. 
    
    \item \emph{Taxonomy Implementation:} we architect an application programming interface and extensible toolkit that realizes the taxonomy in software. This effort is non-trivial given the diversity of methods.  We have released the toolkit into the open source community under the name AI Explainability 360 (AIX360). It is the most comprehensive explainability toolkit across different ways of explaining (see Table~\ref{tab:compare} and Section~\ref{sec:relWork}). 
    
    \item \emph{Algorithmic Enhancements:}  we take several state-of-the-art interpretability methods from the literature and further develop them algorithmically to make them more appropriate and consumable in practical data science applications.  
    
    \item \emph{Educational Material:}  we develop demonstrations, tutorials, and other educational material to make the concepts of interpretability and explainability accessible to non-technical stakeholders.  The tutorials cover several problem domains, including lending, health care, and human capital management, and provide insights into their respective datasets and prediction tasks in addition to their more general educational value.
\end{itemize}

The current version of the toolkit contains eight explainability algorithms: 
\begin{itemize}
    \item \emph{Boolean Decision Rules via Column Generation (BRCG) \cite{BDR}:} Learns a small, interpretable Boolean rule in disjunctive normal form (DNF) for binary classification.
    \item \emph{Generalized Linear Rule Models (GLRM) \cite{GLRM}:} Learns a linear combination of conjunctions for real-valued regression through a generalized linear model (GLM) link function (e.g., identity, logit). 
    \item \emph{ProtoDash \cite{proto}:} Selects diverse and representative samples that summarize a dataset or explain a test instance. Non-negative importance weights are also learned for each of the selected samples.
    \item \emph{ProfWeight \cite{ProfWeight}:} Learns a reweighting of the training set based on a given interpretable model and a high-performing complex neural network. Retraining of the interpretable model on this reweighted training set is likely to improve the performance of the interpretable model.
    \item \emph{Teaching Explanations for Decisions (TED) \cite{TED}:} Learns a predictive model based not only on input-output labels but also on user-provided explanations. For an unseen test instance both a label and explanation are returned.
    \item \emph{Contrastive Explanations Method (CEM) \cite{CEM}:} Generates a local explanation in terms of what is minimally sufficient to maintain the original classification, and also what should be necessarily absent.
    \item \emph{Contrastive Explanations Method with Monotonic Attribute Functions (CEM-MAF) \cite{CEM-MAF}:} For complex images, creates contrastive explanations like CEM above but based on high-level semantically meaningful attributes.
    \item \emph{Disentangled Inferred Prior Variational Autoencoder (DIP-VAE) \cite{DIP-VAE}:} Learns high-level independent features from images that possibly have semantic interpretation.
\end{itemize}
The toolkit also includes two metrics from the explainability literature: Faithfulness \cite{selfEx} and Monotonicity \cite{CEM-MAF}. 

The rest of the paper is organized as follows. In Section \ref{sec:taxonomy}, we describe our taxonomy and show that it is simple to use and comprehensive. In Section \ref{sec:implement}, we describe its software implementation. In Section \ref{sec:enhance}, we outline enhancements made to the algorithms to make them more consumable, along with describing educational material aimed at making the toolkit more accessible to non-AI experts. In Section \ref{sec:relWork}, we discuss other related toolkits and survey papers. In Section \ref{disc}, we summarize our contributions and highlight promising directions along which the toolkit can be further enhanced and extended.

\begin{figure*}[htbp]
  \begin{center}
    \begin{tabular}{c}
      \includegraphics[width=0.8\textwidth]{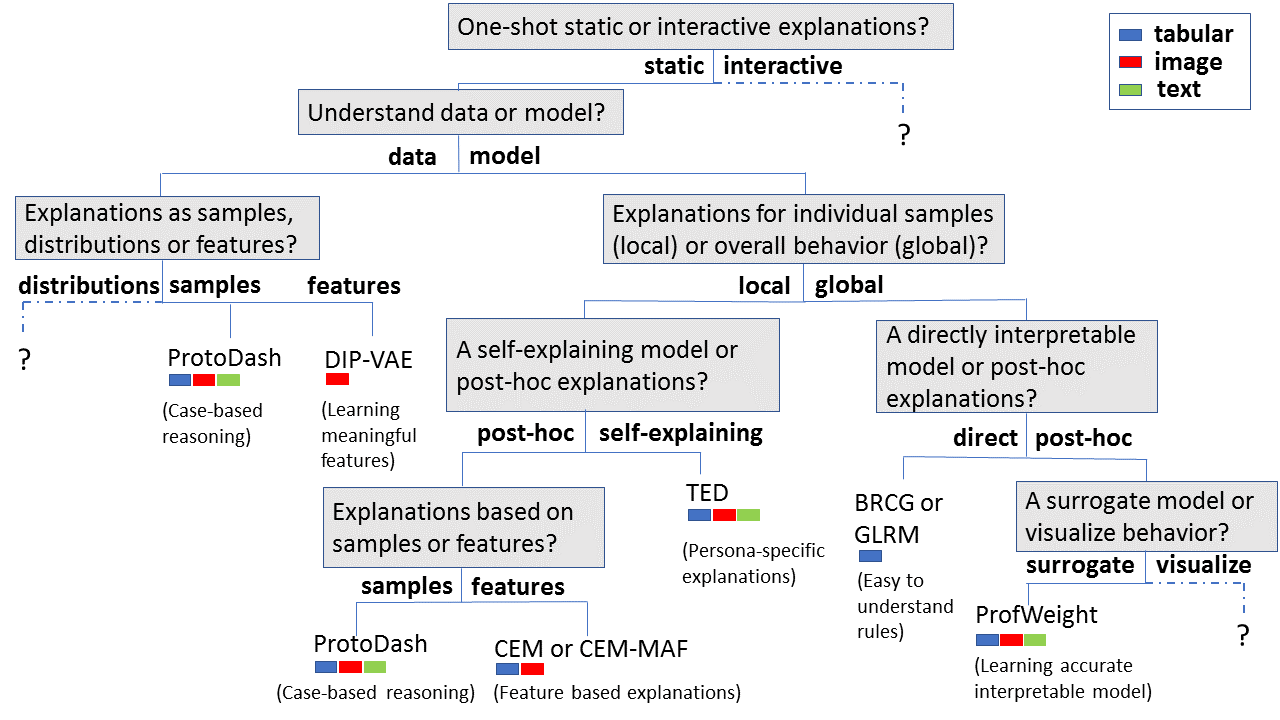}
     \end{tabular}
  \end{center}
  \caption{The proposed taxonomy based on questions about what is explained (e.g.,~data or model), how it is explained (e.g.,~direct/post-hoc, static/interactive) and at what level (i.e.~local/global). The decision tree leaves indicate the methods currently available through the toolkit along with the modalities they can work with. `?' indicates that the toolkit currently does not have a method available in the particular category.}
  \label{fig:algo-guidance}
\end{figure*}

\section{Taxonomy for AI Explainability}
\label{sec:taxonomy}

%\textcolor{red}{[Amit: Ready to be read/edited. ]}

As mentioned in the introduction, explainable artificial intelligence is a rapidly growing research field that has produced many varied methods. Even within the AIX360 toolkit, which represents a small fraction of this literature, different algorithms can cater to different types of consumers (e.g.,~data scientist, loan applicant) who may desire different kinds of explanations (e.g.,~feature-based, instance-based, language-based). %Given these and many more such explainability algorithms, 
It is important therefore to understand the diverse forms of explanations that are available and the questions that each can address. A proper structuring of the explanation space can have large bearing not just on researchers who design and build new algorithms, but perhaps more importantly on practitioners who want to understand which algorithms might be most suitable for their application and might be lost otherwise.

In Figure \ref{fig:algo-guidance}, we provide one such structuring of the explainability space. The structure is in the form of a small decision tree (which is itself considered to be interpretable \cite{freitas2014comprehensible}). Each node in the tree poses a question to the consumer about the type of explanation required. For the majority of consumers who are not experts in AI explainability, the AIX360 website provides a glossary to define terms used in the taxonomy and a guidance document for further context. Herein we note the following definitions:
\begin{itemize}
    \item A \emph{static} explanation refers to one that does not change in response to feedback from the consumer, while an \emph{interactive} explanation allows consumers to drill down or ask for different types of explanations (e.g., through dialog) until they are satisfied. AIX360 focuses on static explanations, like much of the literature to date.
    \item A \emph{local} explanation is for a single prediction, whereas a \emph{global} explanation describes the behavior of the entire model.
    \item A \emph{directly interpretable} model is one that by its intrinsic transparent nature is understandable by most consumers (e.g.,~a small decision tree), whereas a \emph{post-hoc} explanation involves an auxiliary method to explain a model after it has been trained. One could also have a self-explaining model that itself generates local explanations, but may not necessarily be directly interpretable (e.g., rationale generation in text).
    \item A \emph{surrogate} model is a second, usually directly interpretable model that approximates a more complex model, while a \emph{visualization} of a model may focus on parts of it and is not itself a full-fledged model.
\end{itemize}
The leaves in the decision tree are labeled with algorithms in the AIX360 toolkit or a question mark (`?') if the category is not represented in the current toolkit. We also depict the different modalities\footnote{For text input it is assumed that the text has already been embedded in an appropriate vector space.} for which each algorithm could be used. %We argue that this structuring is highly intuitive, and makes it easy for different users to navigate the space and find the right algorithm or class of algorithms appropriate for their task.

\emph{Our primary intention here is to propose a taxonomy that is simple yet comprehensive enough so that it is useful for different types of users. We by no means claim it to be perfect or complete}.

\subsection{Navigating the Taxonomy}

We show that the taxonomy in Figure~\ref{fig:algo-guidance} allows different consumers to navigate the space of explanations and find the right algorithm or class of algorithms appropriate for their task. For simplicity, we narrate the scenario as if the consumers themselves were using the taxonomy, although the more common case is for a data scientist to make these decisions with the consumer's needs in mind.

We
consider a personal finance application where customers are applying for a loan from a financial institution such as a bank. As is typically the case in many deployed applications \cite{procurement}, we assume here that the bank has a complex existing process based on a combination of data-driven approaches and business rules to arrive at the final loan approval decision, which they would like to preserve as much as possible. One consumer persona working within this process is a loan officer, who may wish to validate the approval or denial recommendations of an AI model. Another persona clearly interested in the reasoning behind the decisions is an applicant whose loan was denied. They may want to know about a few factors that could be changed to improve their profile for possible approval in the future. A third persona who might be interested in the overall reasoning of the model is a data science executive. They may want some assurance that in most cases the recommendations made by the model are reasonable.

We now look more closely at these three personas and show how our taxonomy can help them in choosing the right methods.

\begin{itemize}
    \item \emph{Loan Officer:} As mentioned above, the loan officer is interested in validating whether recommendations given by the model for different loan applications are justified. One way for the loan officer to gain confidence in the recommendation made for an individual, whom we will call Alice and whose loan was approved, is to look at other individuals who were similar to Alice and were also approved. Using our taxonomy in Figure \ref{fig:algo-guidance}, they would first go down the ``static'' branch as a single explanation is desired. This would be followed by taking the ``model'', ``local'' and ``post-hoc'' branches as the loan officer wants explanations for particular individuals such as Alice based on an already built model. They would then choose a method under ``samples'' to find other individuals similar to Alice in the dataset. Given the current version of the AIX360 toolkit, they would then choose ProtoDash.
    
    If the training data for the model includes not only past loan decisions but also explanations written by past loan officers, 
    %for all (or at least many) of the approved and denied applicants, 
    then it is possible to give the current loan officer an explanation for Alice in the same form and using the same language. In this case, the loan officer would again traverse the hierarchy in similar fashion as before, except they would now choose the branch ``self-explaining'' and pick TED if using the current version of AIX360.
    
    \item \emph{Applicant:} A different applicant, say Bob, may want to know why his loan was rejected and more importantly what he could do to change the decision in his favor in the future. To support Bob's desire, one would go down the ``static,'' ``model'', ``local'', and ``post-hoc'' branches as he wants an explanation of the model's decision just for himself. One would then choose an appropriate method under ``features'' as Bob wants an explanation based on his profile. Given the current version of AIX360, CEM would be chosen to obtain such an explanation.
    
    \item \emph{Bank Executive:} Bank executives may not care so much about specific applicants as about the overall behavior of the model and whether it is, in general, basing its recommendations on robust justifiable factors. In this case, executives would follow the ``static'' and ``model'' branches followed by ``global'' as they want a high-level understanding of the model. Since the model is already deployed, they would then follow the ``post-hoc'' branch and finally choose ``surrogate'' to obtain an interpretation of the original model using another model they can understand. In this case, a natural choice from our toolkit would be ProfWeight. However, 
    %one should note that
    a directly interpretable method such as BRCG could also be used here where it is trained on the predictions of the original complex model. This follows the common strategy of model compression \cite{modelcompr} or knowledge distillation \cite{distill}.%, which are prevalent in machine learning.
\end{itemize}

\subsection{Categorizing Common Classes of Explainability Methods}
\label{sec:taxonomy:categorizing}
We now show how some popular classes of explainability methods can be effectively categorized in our taxonomy. Again, we do not claim that this is a complete listing, although most of the methods reviewed by Guidotti et al.~\cite{guidotti2018survey} are included.

\begin{itemize}
    \item \emph{Saliency Methods [static $\rightarrow$ model $\rightarrow$ local
    $\rightarrow$ post-hoc
    $\rightarrow$ features]:} Saliency based methods \cite{saliency,selvaraju2016grad}, which highlight different portions in an image whose classification we want to understand, can be categorized as local explanation methods that are static and provide feature-based explanations in terms of highlighting pixels/superpixels. In fact, popular methods such as LIME \cite{lime} and SHAP \cite{unifiedPI} also fall under this umbrella. Counterfactual explanations \cite{counterfactual}, which are similar to contrastive explanations where one tries to find a minimal change that would alter the decision of the classifier, are another type of explanation in this category.
    
    \item \emph{Neural Network Visualization Methods [static $\rightarrow$ model $\rightarrow$ global $\rightarrow$ post-hoc $\rightarrow$ visualize]:} Methods that visualize intermediate representations/layers of a neural network \cite{nguyen2016multifaceted,deepling} would fall under the global post-hoc category as people typically use these visualizations to gain confidence in the model and inspect the type of high-level features being extracted.
    
    \item \emph{Feature Relevance Methods [static $\rightarrow$ model $\rightarrow$ global $\rightarrow$ post-hoc $\rightarrow$ visualize]:} Methods such as partial dependence plots (PDP) and sensitivity analysis \cite{molnarbook} are used to study the (global) effects of different input features on the output values and are typically consumed through appropriate visualizations (e.g., PDP plots, scatter plots, control charts).
    
    \item \emph{Exemplar Methods [static $\rightarrow$ model $\rightarrow$ local $\rightarrow$ post-hoc $\rightarrow$ samples]:} Methods to explain the predictions of test instances based on similar or influential training instances \cite{l2c,infl,proto} would be considered as local explanations in the form of samples.
    
    \item \emph{Knowledge Distillation Methods [static $\rightarrow$ model $\rightarrow$ global $\rightarrow$ post-hoc $\rightarrow$ surrogate]:} Knowledge distillation-type approaches \cite{distill,modelcompr,modelcompr2}, which learn a simpler model based on a complex model's predictions, would be considered as global interpretations that are learned using a post-hoc surrogate model.
    
    \item \emph{High-Level Feature Learning Methods [static $\rightarrow$ data $\rightarrow$ features]:} Methods that learn high-level features in an unsupervised manner \cite{infogan,DIP-VAE} through variational autoencoder or generative adversarial network frameworks would naturally fall under the data followed by features category. Even supervised methods \cite{tcav} of learning high-level interpretable features would lie in this category.
    
    \item \emph{Methods that Provide Rationales [static $\rightarrow$ model $\rightarrow$ local
    $\rightarrow$ self]:} Work in the natural language processing and computer vision domains that generates rationales/explanations derived from input text \cite{lei2016rationalizing,Yessenalina:2010:AGA:1858842.1858904,hendricks-2016} would be considered as local self explanations. Here however, new words or phrases could be generated so the feature space can be richer than the original input space.
    
    \item \emph{Restricted Neural Network Architectures [static $\rightarrow$ model $\rightarrow$ global $\rightarrow$ direct]:} Methods that propose certain restrictions on the neural network architecture \cite{selfEx,interpNN} to make it interpretable, yet maintain richness of the hypothesis space to model complicated decision boundaries would fall under the global directly interpretable category.
\end{itemize}

% \begin{center}
% \begin{tabular}{ |c|c|c|c|c|c|c| } 
%  \hline
%  Toolkit & Data & Directly & Local & Global & Persona-Specific & Metrics\\
%  & Explanations & Interpretable & post-hoc & post-hoc & Explanations & \\
%  \hline
% \end{tabular}
% \end{center}

\begin{figure*}[ht]
  \centering
      \includegraphics[width=0.8\textwidth]{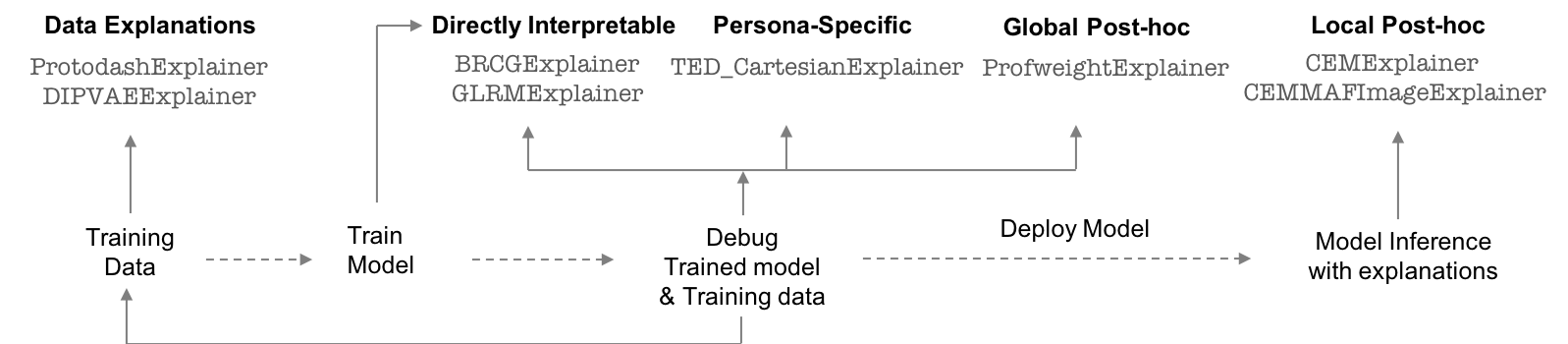}
  \caption{Organization of AIX360 explainer classes according to their use in various steps of the AI modeling pipeline.}
  \label{fig:taxonomy-implementation}
\end{figure*}

\section{Implementation of Taxonomy}
\label{sec:implement}
%Implementation of the taxonomy (software architecture/design--class hierarchy, extensibility, ease of use) in relation to above.
%a) Non-experts
%b) Developers/data scientists

The AIX360 toolkit aims to provide a unified, flexible, and easy to use programming interface and an associated software architecture to accommodate the diversity of explainability techniques required by various stakeholders. %in the AI value chain,  that 
The goal is to be amenable both to data scientist users, who may not be experts in explainability, as well as algorithm developers. Toward this end, we make use of a %Non-expert users and data scientists are exposed to an intuitive 
programming interface that is similar to popular Python model development tools (e.g.,~scikit-learn) and construct a hierarchy of Python classes corresponding to explainers for data, models, and predictions. Algorithm developers can inherit from a family of base class explainers in order to integrate new explainability algorithms. 

The organization of explainer classes takes into account not only the taxonomy described in Section~\ref{sec:taxonomy} but also the perspective of the primary users of the toolkit, data scientists and developers. As such, it is not identical to the taxonomy in Figure~\ref{fig:algo-guidance} although there are clear correspondences. Instead, we have organized the base class explainers according to the AI modeling pipeline shown in Figure \ref{fig:taxonomy-implementation}, based upon their use in offering explanations at different stages. Details are as follows:

\begin{itemize}
\item \emph{Data explainers:}  These explainers are implemented using the base class DIExplainer (Directly Interpretable\footnote{In the codebase, which predates the taxonomy in Figure~\ref{fig:algo-guidance}, the term ``directly interpretable'' just refers to the absence of a surrogate model or method.} unsupervised Explainer), which provides abstract methods for explainers that use unsupervised techniques to explain datasets. The AIX360 explainers that inherit from this class include ProtodashExplainer and DIPVAEExplainer. 

\item \emph{Directly interpretable explainers:} These explainers are implemented using the base class DISExplainer (Directly Interpretable Supervised Explainer), which supports explainers that train interpretable models directly from labelled data. The AIX360 explainers that inherit from this class and implement its methods include BRCGExplainer and GLRMExplainer. Additionally, the TED\_CartesianExplainer, which can train models based on data labelled with persona-specific explanations, also inherits from DISExplainer. Listing~\ref{lst:dise} shows a subset of abstract methods provided by DISExplainer and an example illustrating the use of BRCGExplainer.  
 
\item \emph{Local post-hoc explainers:} These explainers are further sub-divided into black-box and white-box explainers. The black-box explainers are model-agnostic and generally require access only to a model's prediction function while the white-box explainers generally require access to a model's internals such as its loss function. Both classes of explainers are implemented in AIX360 via base classes LocalBBExplainer and LocalWBExplainer, respectively. The AIX360 explainers CEMExplainer and CEMMAFExplainer both inherit from LocalWBExplainer. Listing \ref{lst:dise}c shows an example of using the CEMExplainer to obtain pertinent negatives corresponding to MNIST images.

\item \emph{Global post-hoc explainers:} These explainers are sub-divided into black-box and white-box explainers as above. The GlobalWBExplainer base class includes abstract methods for explainers that train a more interpretable surrogate model given an original source model along with its data. The ProfweightExplainer in AIX360 inherits from this base class. 
\end{itemize}
%  \vspace{-3mm}

\begin{tiny}
\begin{lstlisting}[caption=(a) DISExplainer base class; (b) Example illustrating the use of BRCGExplainer (directly interpretable supervised); (c) Example illustrating the use of CEMExplainer (local white-box), label={lst:dise}, language=Python ] 
                         (a)
class DISExplainer(ABC):
    def __init__(self, *argv, **kwargs):
     #Initialize a Directly Interpretable Supervised Explainer object.
                
    @abc.abstractmethod
    def fit(self, *argv, **kwargs):
        #Fit an interpretable model on data.
        raise NotImplementedError  #only if not implemented by inheriting class

    @abc.abstractmethod
    def explain(self, *argv, **kwargs):
        #Explain the model
        raise NotImplementedError     
    ...
                          (b)      
from aix360.algorithms.rbm import BRCGExplainer, BooleanRuleCG
from aix360.datasets import HELOCDataset
# load the FICO Heloc dataset
(_, x_train, _, y_train, _) = HELOCDataset().split()
# Instantiate a directly interpretable explainer in conjunctive normal form
br = BRCGExplainer(BooleanRuleCG(CNF=True))
# train the interpretable model on data
br.fit(x_train, y_train)
# print the CNF rules
print (br.explain()['rules'])

                          (c)
from aix360.algorithms.contrastive import CEMExplainer
from aix360.datasets import MNISTDataset
# load normalized MNIST data 
data = MNISTDataset()
# Instantiate a local post-hoc explainer
explainer = CEMExplainer(mnist_model)
# chose an input image
input_image = np.expand_dims(data.test_data[0], axis=0)
# obtain pertinent negative explanations for an image
(pn_image, _, _) = explainer.explain_instance(input_image, arg_mode='PN") 
\end{lstlisting}
\end{tiny}

In addition to the above base classes, AIX360 includes dataset classes to facilitate loading and processing of commonly used datasets so that users can easily experiment with the implemented algorithms.

\section{Enhancements to Bring Research Innovations to Consumers}
\label{sec:enhance}

%\textcolor{red}{[DW: I think all of Section~\ref{sec:enhance} is ready to be reviewed, including Section~\ref{sec:enhance:demo}, even if Mike may add something there.]}

Beyond the higher-level contributions discussed in Sections~\ref{sec:taxonomy} and \ref{sec:implement}, AIX360 also includes enhancements to individual explainability methods as well as complementary components. %consists of much more than a package of implementations of explainability methods as described in papers. 
These enhancements take methods as described in papers, which may only benefit machine learning researchers and expert data scientists, and make them accessible to a broader range of consumers. 

The enhancements presented in this section run the gamut from algorithmic to educational. For BRCG, a simplified algorithm was developed as detailed in Section~\ref{sec:enhance:BRCG} to avoid the need for an integer programming solver, which would be a major barrier to adoption. For TED, the barrier is the lack of datasets containing explanations along with labels. Section~\ref{sec:enhance:TED} discusses the synthesis of such a dataset based on rules for an employee retention application. In Sections~\ref{sec:enhance:GLRM} and \ref{sec:enhance:ProtoDashCEM}, we discuss extensions that are built on top of existing model/algorithm outputs to improve their consumability: visualization for GLRM, and feature importance for ProtoDash and CEM. Section~\ref{sec:enhance:metrics} documents the explainability metrics included in AIX360. Lastly, Section~\ref{sec:enhance:demo} describes an interactive web demo targeted at non-experts and featuring a credit approval use case approached from the perspectives of three types of consumers. Also discussed in Section~\ref{sec:enhance:demo} are tutorials aimed at data scientists looking to introduce explainability into different application domains.

\subsection{Boolean Decision Rules via Column Generation: ``Light'' version}
\label{sec:enhance:BRCG}

The BRCG algorithm produces a binary classifier in the form of a disjunctive normal form (DNF) rule (conjunctive normal form (CNF) is also possible through a simple transformation). The published version of BRCG \cite{BDR} relies on the use of an integer programming solver such as CPLEX, which is often prohibitive due to cost or other barriers to access. In AIX360, we have replaced the integer programming components with a heuristic beam search algorithm and have modified the problem formulation accordingly. We call the resulting algorithm a ``light'' version of BRCG (BRCG-light). 

To describe BRCG-light, we recall some notation from \cite{BDR} (please see \cite{BDR} for more context in general). Let $n$ denote the training sample size and $\cP$ and $\cZ$ denote the sets of indices corresponding to positively and negatively labeled samples. Let $\cK$ index the set of all possible (exponentially many) clauses that can be included in the DNF rule. These clauses are conjunctions of the features, which have been binarized as discussed in \cite{BDR}. The binary-valued decision variable $w_k$ indicates whether clause $k \in \cK$ is included in the rule, and $c_k$ is the complexity of the clause. For sample $i$, $\cK_i \subseteq \cK$ is the subset of clauses satisfied by the sample (which evaluate to true), and for $i \in \cP$, $\xi_i$ indicates a false negative error. 

BRCG-light attempts to solve the following problem related to \cite[eq.~(1)-(4)]{BDR}:
\begin{equation}\label{eqn:BRCGmaster}
    \begin{split}
        \min_{w, \xi} \quad &\frac{1}{n} \sum_{i\in\cP} \xi_i + \frac{1}{n} \sum_{i\in\cZ} \sum_{k\in\cK_i} w_k + \sum_{k\in\cK} c_k w_k\\
        \text{s.t.} \quad &\xi_i + \sum_{k\in\cK_i} w_k \geq 1, \quad \xi_i \geq 0, \quad i \in \cP\\
        &w_k \in \{0, 1\}, \quad k \in \cK.
    \end{split}
\end{equation}
Like in \cite{BDR}, the first two terms of the objective function in \eqref{eqn:BRCGmaster} represent the Hamming loss of the classifier. Different from \cite{BDR}, the last term in the objective is a penalty on the complexity of the rule, whereas \cite{BDR} uses an upper bound on complexity. Herein the clause complexities $c_k$ are parametrized by a fixed cost $\lambda_0$ plus a variable cost $\lambda_1$ multiplying the number of conditions (literals) in the clause (see the objective function in \eqref{eqn:BRCGpricing} below), whereas \cite{BDR} only considered the equivalent of $\lambda_1 = \lambda_0$. These modifications bring BRCG-light closer to GLRM \cite{GLRM} in two respects: 1) They conform to the GLRM problem formulation in which rule complexity is penalized in the same way, giving AIX360 users a common interface; 2) They allow reuse of a GLRM sub-algorithm as explained below. A minor difference from \cite{BDR} is that the first two terms in the objective in \eqref{eqn:BRCGmaster} are scaled by $1/n$ to facilitate the setting of $\lambda_0$ and $\lambda_1$ across datasets of different sizes. 

BRCG-light uses the same column generation approach as in \cite{BDR} to effectively handle the exponential number of variables in \eqref{eqn:BRCGmaster}. Below we summarize the steps in this approach:
\begin{enumerate}
    \item Solve the linear programming (LP) relaxation of \eqref{eqn:BRCGmaster}, obtained by relaxing the $w_k \in \{0,1\}$ constraint to $w_k \geq 0$, restricted to a small subset of clauses $\cS \subset \cK$ in place of $\cK$.\label{step:restrictedLP}
    \item Solve a \emph{pricing problem} given below in \eqref{eqn:BRCGpricing} to find clauses omitted from step \ref{step:restrictedLP} that can improve the current LP solution and add them to $\cS$.\label{step:pricing}
    \item Repeat steps \ref{step:restrictedLP} and \ref{step:pricing} until no improving clauses can be found.
    \item Solve the unrelaxed version of \eqref{eqn:BRCGmaster} restricted to the final subset $\cS$.\label{step:restrictedIP}
\end{enumerate}
The pricing problem for BRCG-light is also slightly modified from that in \cite{BDR}. Define $\mu_i$, $i \in \cP$ to be the optimal dual variables associated with the constraints $\xi_i + \sum_{k\in\cK_i} w_k \geq 1$ in \eqref{eqn:BRCGmaster}, whose values are determined by solving the LP in step \ref{step:restrictedLP} above. Denote by $x_{ij}$, $i=1,\dots,n$, $j=1,\dots,d$, the value of the $i$th sample of the $j$th binarized feature, and let $\bar{x}_{ij} = 1 - x_{ij}$. Clauses are encoded by a binary-valued vector $z$ with components $z_j$ indicating whether the $j$th feature participates in the clause. The BRCG-light pricing problem is as follows:
\begin{equation}\label{eqn:BRCGpricing}
    \begin{split}
        \min_{z, \delta} \quad &\frac{1}{n} \sum_{i\in\cZ} \delta_i - \sum_{i\in\cP} \mu_i \delta_i + \lambda_0 + \lambda_1 \sum_{j=1}^d z_j\\
        \text{s.t.} \quad &\delta_i + \sum_{j=1}^d \bar{x}_{ij} z_j \geq 1, \quad \delta_i \geq 0, \quad i \in \cZ\\
        &\delta_i + z_j \leq 1, \quad i \in \cP, \quad j: \bar{x}_{ij} = 1\\
        &\sum_{j=1}^d z_j \leq D\\
        &z_j \in \{0, 1\}, \quad j = 1, \dots, d.
    \end{split}
\end{equation}

In \cite{BDR}, steps \ref{step:pricing} and \ref{step:restrictedIP} of the column generation procedure are performed by calling the integer programming solver CPLEX. For BRCG-light, we avoid integer programming in step \ref{step:pricing} by observing instead that \eqref{eqn:BRCGpricing} is an instance of the column generation subproblem for GLRM \cite[eq.~(9)]{GLRM}, with the following identifications: $r_i = 1$ for $i \in \cZ$ (corresponding to set $\cI_+$ in \cite{GLRM}), $r_i = -n \mu_i$ for $i \in \cP$ (corresponding to set $\cI_-$), and $a = \delta$. As a consequence, we may use the heuristic beam search algorithm described in \cite{GLRM} to obtain a good feasible solution to \eqref{eqn:BRCGpricing}. The constraint $\sum_{j} z_j \leq D$ in \eqref{eqn:BRCGpricing} can be met by setting the maximum degree (i.e.~depth) of the beam search to $D$. Since the column generation procedure requires only a solution to \eqref{eqn:BRCGpricing} with negative objective value to improve the LP solution from step \ref{step:restrictedLP}, and not necessarily an optimal solution to \eqref{eqn:BRCGpricing}, the substitution of a heuristic algorithm for integer programming has less of an effect. On the other hand, if the beam search algorithm fails to find a solution to \eqref{eqn:BRCGpricing} with negative objective value, then we do not have a certificate that such a solution does not exist, unlike with integer programming run to completion.

BRCG-light also avoids integer programming in step \ref{step:restrictedIP}. For this, we observe that \eqref{eqn:BRCGmaster} (with or without the restriction to subset $\cS$) is in fact an instance of a slight generalization of the GLRM column generation subproblem \cite[eq.~(9)]{GLRM}. The identifications are as follows: $r_i = 1$ for $i \in \cP$ corresponding to set $\cI_+$, an empty set $\cI_- = \emptyset$, $a = \xi$, $\lambda_0 = 0$, indices $k \in \cS$ instead of $j = \{1,\dots,d\}$, and a binary variable $b_{ik}$ indicating whether the $i$th sample satisfies the $k$th clause in place of $\bar{x}_{ij}$. The slight generalization is that the sum $\lambda_1 \sum_{j=1}^d z_j$ in \cite[eq.~(9)]{GLRM} is replaced by the weighted sum $\sum_{k\in\cS} \lambda_k z_k$ with weights $\lambda_k = c_k + (1/n) \sum_{i\in\cZ} b_{ik}$. It follows that a slight generalization of the beam search algorithm in \cite{GLRM} can produce good solutions to \eqref{eqn:BRCGmaster} restricted to $\cS$, as required in step \ref{step:restrictedIP}.

In terms of implementation in AIX360, the beam search algorithm from \cite{GLRM} has been generalized as discussed above and made available to both BRCG-light and GLRM as a common Python module. The modules implementing GLRM follow \cite{GLRM} plus the enhancement of Section~\ref{sec:enhance:GLRM} below while the module implementing BRCG-light is novel.

In 
\ifappendix
Appendix~\ref{sec:BRCGexpt}, 
\else
the supplemental material,
\fi
we present trade-off curves between classification accuracy and rule complexity for BRCG-light in comparison to the integer programming version of BRCG (BRCG-IP) as well as previous rule learning methods evaluated in \cite{BDR}. These comparisons show that BRCG-light occupies an intermediate position, inferior to BRCG-IP as may be expected, but superior to the earlier methods. In particular, a beam search width of $5$ (the default in AIX360) gives better results than smaller widths on most datasets. Larger widths might give still better results with increased computational complexity. 

\subsection{Teaching Explanations for Decisions: Data synthesis}
\label{sec:enhance:TED}

The TED framework~\cite{TED} requires a training dataset that includes
explanations ($E$) in addition to features ($X$) and labels ($Y$).  These
explanations are used to teach the framework what are appropriate
explanations for a given feature set in the same manner that a typical
training dataset teaches what are appropriate labels ($Y$).  To address
this gap between the needs of the framework (datasets with explanations) and current practice (datasets with no explanations), we
devised a general strategy for creating synthetic datasets, modeled
after the approach described in~\cite{TED} for predicting tic-tac-toe moves.

Consider the use case of determining which employees should be
targeted with a retention action to dissuade them from leaving the company \cite{Singh2012}.
Ideally, we would ask a human resource (HR) specialist with deep knowledge
of the circumstances for employee retention to develop a training
dataset where each employee, represented as a feature vector, would have a prediction of whether the
employee was at risk to leave, justified with an explanation.

Since we do not have such information, we simulate this process by
encoding the HR specialist's knowledge into a set of rules.  Each
rule is a function with a feature vector as its inputs and the
prediction of whether the employee is a retention risk ($Y$) as its output.
We use the rule number as the explanation for this prediction since it should be meaningful to the 
target persona.

Given these rules, we can then generate a synthetic dataset of any
size by first generating a random collection of feature vectors and
then applying the rules to determine the appropriate predictions and
explanations.  AIX360 contains such a dataset for employee retention
with 10,000 entries, created by an available python script that can be modified
by changing the distribution of features, changing the rules, or the size of the dataset.
It can also serve as a example for the generation of datasets for other classification problems.

In summary, to use this approach for a classification problem, one must determine the following:
\begin{enumerate}
\item the features of the problem
\item the distribution of each feature to be generated and any dependence among features
\item the rules for mapping feature vectors to predictions; the rule number will be used as the explanation
\end{enumerate}

For the employee retention dataset included with AIX360, we used 8
features (Position, Organization, Potential, Rating, Rating Slope,
Salary Competitiveness, Tenure Duration, and Position Duration) and
various distributions for each feature, resulting in a feature space
of over 80 million possibilities.  %(See the employee retention tutorial notebook for more details.)  
We created 25 rules, motivated by
common retention scenarios, such as not getting a promotion in a
while, not being paid competitively, receiving a disappointing
evaluation, being a new employee in  organizations with
inherently high attrition, not having a salary that is consistent with
positive evaluations, mid-career crisis, etc.  In the current
implementation of this use case, we further vary the application of these
rules depending on various positions and organizations. For example, an organization may have %in our fictitious company organization \#1 has 
much higher attrition
because their skills are more transferable outside the company.

This dataset reinforces %illustrates 
two advantages of the TED framework:
\begin{itemize}
\item the explanations match the terminology of the domain; they represent higher-level semantic concepts, e.g.,~``not being paid competitively'', created by the domain expert, rather than mathematical combinations of feature values,

\item a single explanation ``not being paid competitively'' can occur in multiple ways in the feature space.  ``Not being paid competitively'' rules will likely have different flavors depending on one's organization, position, and tenure.  Although this will map to multiple rules, the explanation provided to the consumer can be a single concept.  %The retention tutorial notebook in AIX360 provides an example.

\end{itemize}

\subsection{Generalized Linear Rule Models: Visualization}
\label{sec:enhance:GLRM}

The GLRM algorithm produces linear combinations of conjunctive `AND' rules, which provide predictions directly in the case of linear regression or through a transformation in the case of generalized linear models (e.g.,~logistic regression). GLRM models can therefore be interpreted like linear models and their constituent rules are also readily understood individually. Nevertheless, the discussion in \cite[Sec.~6]{GLRM} as well as our experience in putting together the AIX360 web demo and tutorials (Section~\ref{sec:enhance:demo}) raise the question of the degree to which a GLRM model with (say) tens of rules is interpretable. While interpretation is still feasible, it may require a non-trivial investment of effort. 

The authors of \cite{GLRM} suggest a decomposition of GLRM models to reduce the effort of interpretation. Since first-degree rules, i.e.~those that place a single condition on one feature, do not create interactions between features, they collectively form a type of generalized additive model (GAM) \cite{hastie1990GAM}, i.e.~a sum of univariate functions $f_j$ of individual features $x_j$,
\begin{equation}\label{eqn:GAM}
    f(x) = \sum_{j=1}^d f_j(x_j).
\end{equation}
The same is true of linear functions of original, unbinarized numerical features (so-called ``linear terms'') if they are included in the GLRM model, so these can be added to the GAM. The benefit of having a GAM is that it can be visualized as a series of plots of the functions $f_j$, thus summarizing the effects of all first-degree rules and linear terms. Higher-degree rules, which do involve feature interactions, can then be studied separately.

In AIX360, we have made the above suggestion a reality by equipping GLRM models with the ability to visualize the GAM part of the model. This is implemented as a \texttt{visualize()} method of the GLRM Python classes. Given a trained GLRM object and for each original feature $x_j$, the \texttt{visualize()} method collects the associated first-degree rules and linear terms along with their model coefficients. It also collects information needed for plotting such as the domain of $x_j$ and its mean and standard deviation if they were removed during standardization. In addition, the method computes the importance of each feature to the GAM (though not necessarily to the overall GLRM), which is used to order the plots and possibly for other purposes. Importance is measured by the variance $\mathrm{var}(f_j(X_j))$ with respect to the empirical distribution of $X_j$. This accounts for both the variation of $f_j$ as well as that of $X_j$ (a near-constant function $f_j$ or near-constant values for $X_j$ would both yield low variance). Lastly, the \texttt{explain()} method of the GLRM classes, which prints a list of rules and linear terms, has the option of printing only higher-degree rules.

\subsection{ProtoDash and Contrastive Explanations: Feature importance}
\label{sec:enhance:ProtoDashCEM}

%\textcolor{red}{[Amit: Ready to be read/edited. ]}

The base algorithms for both ProtoDash \cite{proto} and CEM \cite{CEM} do not return feature importances. What they return are vectors that are the same dimension as the input. This might be sufficient for an expert data scientist to use to interpret, however, it might still pose a challenge to a less experienced user. In the AIX360 toolkit we add another layer of interpretability by also discerning features that are important in more easily interpreting the output. This is clearly observed in the tutorials that showcase the utility of these methods.

In particular, let $\bx$ denote the original input with $x_{[j]}$ denoting its $j^{\textrm{th}}$ dimension. For ProtoDash when used to explain an input, let $\bx'$ denote a prototype found by it for $\bx$. Then we compute the feature importance for feature $j$ relative to $\bx'$ denoted by $\theta_j$, where $\sigma_j$ denotes the standard deviation of feature $j$ as follows:
\begin{equation}
\label{eqn:proto}
\theta_j = \exp{\frac{-~\left|x_{[j]}-x'_{[j]}\right|}{\sigma_j}}
\end{equation}
Here $|\cdot|$ denotes absolute value. The above equation creates a feature importance value in $[0,1]$. The higher the value more similar the feature and hence higher its importance. The top $\theta_j$s could inform the user of which and how many features of the prototype $\bx'$ are close to that of $\bx$ making it a good exemplar based explanation for it.

For CEM given an input $\bx$ classified by a classifier in class $y$ we output two vectors, namely, a pertinent positive (PP) vector $\bx^{pp}$ and a pertinent negative (PN) vector $\bx^{pn}$. A PP informs us of what minimal values for different features not exceeding those in $\bx$ would be sufficient to obtain the same prediction $y$. While, a PN informs us of what minimal features if added or increased in value would change the classification of $\bx$ to something other than $y$. Given these definitions we compute the feature importances for the $j^{\textrm{th}}$ feature for PPs and PNs as follows:

\begin{equation}
\label{eqn:cem}
\begin{split}
\theta^{pp}_j &=1- \exp{\frac{-\left|x^{pp}_{[j]}\right|}{\sigma_j}}\\
\theta^{pn}_j &=1- \exp{\frac{-\left|x_{[j]}-x^{pn}_{[j]}\right|}{\sigma_j}}
\end{split}
\end{equation}
The higher the corresponding $\theta_j$ values, the more important that feature is to either maintain or change the class depending on if its a PP or a PN respectively. The feature importances thus provide insight to the user on which factors are possibly most critical in the classification, which we expose now through the toolkit.

\subsection{Explainability metrics}
\label{sec:enhance:metrics}

%\textcolor{red}{[Amit: Ready to be read/edited. ]}

Although explanations are difficult to quantitatively evaluate and the gold standard is typically human evaluation \cite{rsi}, the AIX360 toolkit provides two quantitative metrics that serve as proxies of how ``good'' a particular feature-based local explanation is likely to be.

The first metric we implement is a version of the \emph{Faithfulness} metric \cite{selfEx}, which tries to evaluate if the feature importances returned by an explainability method are the correct ones. The method starts by replacing the most important feature value by another value termed as the base value which the user supplies as being a no information or no-op value. For grey-scaled images the base value would be zero for pixels, indicating that it is blank. Once this replacement has been done the new example is passed to the classifier and its prediction probability for the original predicted class is noted. The most important feature value is then restored and the second most important features value is replaced by the base value and again the prediction probability is noted. This process continues for all features that we have feature importance for. Given a feature importance vector $\mathbf{\theta}$ and the corresponding prediction probabilities $\mathbf{p}$, with $\rho$ denoting Pearsons correlation we compute the following metric:
\begin{equation}
    \phi=-\rho(\mathbf{\theta},\mathbf{p})
\end{equation}
The higher the $\phi$ the better the explanation. The intuition is that removing more important features should reduce the confidence of the classifier more than removing less important ones.

The second metric is \emph{Monotonicity} \cite{CEM-MAF}, which simply tries to measure if adding more positive evidence increases the probability of classification in the specified class. In this case we start with the base value vector and keep incrementally replacing the features that have positive relevance with their actual values in increasing order of importance and note the classification probabilities in each case. If the features are (independent) positively correlated with the decision, then the class probabilities should monotonically increase.

Although both of these metrics are not novel in themselves, we are possibly the first explainability toolkit that has quantitative metrics to measure the ``goodness'' of explanations, as is indicated in Table \ref{tab:compare}.

\subsection{Web demo, tutorials and resources}
\label{sec:enhance:demo}

AIX360 was developed with the goal of providing accessible resources on explainability to non-technical stakeholders. Therefore, we include multiple educational materials to both introduce the explainability algorithms provided by AIX360, and to demonstrate how different explainability methods can be applied in real-world scenarios. These educational materials include a general guidance for the key concepts of explainability, a web demo that illustrates the usage of different explainability methods, and multiple tutorials.

The web demo was created based on the  FICO Explainable Machine Learning Challenge dataset \cite{FICO2018}, a real-world scenario where a machine learning system is used to support decisions on loan applications by predicting the repayment risk of the applicants. Sample screenshots from the demo are shown in 
\ifappendix
Figure~\ref{fig:demo}.
\else
Figure~6 in the supplemental material.
\fi
The demo highlights that three groups of people -- data scientists, loan officers, and bank customers -- are involved in the scenario and their needs are best served by different explainability methods. For example, although the data scientist may demand a global understanding of model behavior through an interpretable model, which can be provided by the GLRM algorithm, a bank customer would ask for justification for their loan application results, which can be generated by the CEM algorithm. We use storytelling and visual illustrations to guide users of AIX360 through these scenarios of different explainability consumers.

The AIX360 toolkit currently includes five tutorials in the form of Jupyter notebooks that show data scientists and other developers how to use different explanation methods across several application domains. The tutorials thus serve as an educational tool and potential gateway to AI explainability for practitioners in these domains. The tutorials cover:
\begin{itemize}
    \item Using three different methods to explain a credit approval model to three types of consumers, based on the FICO Explainable Machine Learning Challenge dataset \cite{FICO2018}.
    \item Creating directly interpretable healthcare cost prediction models for a care management scenario using Medical Expenditure Panel Survey data \cite{MEPS}.
    \item Exploring dermoscopic image datasets used to train machine learning models that help physicians diagnose skin diseases.
    \item Understanding National Health and Nutrition Examination Survey datasets to support research in epidemiology and health policy.
    \item Explaining predictions of a model that recommends employees for retention actions from a synthesized human resources dataset.
\end{itemize}
Beyond illustrating the application of different methods, the tutorials also provide considerable insight into the datasets that are used and, to the extent that these insights generalize, into the respective problem domains. These insights are a natural consequence of using explainable machine learning and could be of independent interest.

\section{Related Work}
\label{sec:relWork}
%\textcolor{red}{[Related work (put chart comparing to other toolkits, post-hoc proxy model issue, unsupervised methods, metrics).]}

%\textcolor{red}{[Vera: For now I focus on reviewing taxonomy related works, and ignore more high level topics such as motivation and history for XAI. ]}
In this section, we review existing work that relates specifically to the two main topics of this paper: taxonomies and toolkits for explainable AI. We do not focus on the much larger literature on explainability methods themselves, although Section~\ref{sec:taxonomy:categorizing} does connect many of these to our taxonomy. %\textcolor{red}{[DW: Any surveys of explainability methods we can cite here?]} \textcolor{red}{[Vera: Yes. Guidotti et al. is about mapping methods]}

%Explainable artificial intelligence is a rapidly growing research field that has produced many methods and techniques to make AI more transparent. 
The rapid expansion of explainable AI research has motivated several recent papers that survey the field and 
identify criteria to systematically classify explainability methods and techniques~\cite{molnarbook,adadi2018peeking,carvalho2019machine,guidotti2018survey}. For example, Adadi and Berrada~~\cite{adadi2018peeking} conduct an extensive literature review and classify explainability methods according to three criteria: 1) The complexity or interpretability of the model to be explained, i.e., whether it is directly interpretable (\textit{intrinsic}), a category that we also have in Figure~\ref{fig:algo-guidance}, or a black-box model that requires post-hoc methods, also represented in Figure~\ref{fig:algo-guidance}; %applying methods to analyze the model after training (\textit{post-hoc}); 
2) The scope of interpretability, corresponding to our local/global split in Figure~\ref{fig:algo-guidance}; %which differentiates between the needs to understand the entire model behavior (\textit{global}) versus a single prediction (\textit{local}); 
and 3) The dependency on the model used, i.e., whether the explanation technique applies to any type of machine learning model (\textit{general}) or to only one type (\textit{model specific}). Besides these three criteria, Carvalho et al.~\cite{carvalho2019machine} add a criterion on the stage of model development where an explainability method is applied, i.e., whether it is before (\textit{pre-model}), during (\textit{in-model}), or after (\textit{post-model}) building the machine learning model. As discussed in Section~\ref{sec:implement}, the software implementation of our taxonomy is organized around a similar consideration. Lastly, several recent papers advocate for the vision of \textit{interactive} explanation~\cite{miller2018explanation,hohman2019gamut,weld2018intelligible, prospector}, allowing users to drill down or ask for different types of explanation (e.g., through dialogues) until satisfied with their understanding. In Figure~\ref{fig:algo-guidance} we include static/interactive as the first split, and note that there is a scarcity of available techniques for interactive explanations. The proposed taxonomy (Figure~\ref{fig:algo-guidance}) and software architecture (Section~\ref{sec:implement}) add to the criteria above by making both broader distinctions (data vs.~ model) as well as finer distinctions (features vs.~samples, black-box vs.~white-box post-hoc).

Although the above criteria provide a useful understanding of the interpretability literature, they by themselves do not imply %these high-level criteria still lack 
actionability in terms of guiding %for one to navigate the space of explainable AI, for example to support 
a practitioner in selecting explainability methods. A recent paper by Guidotti et al.~\cite{guidotti2018survey} attempts to create a detailed mapping of popular explainability \textit{methods} (explanator) such as feature importance, saliency masks, and prototype selection to the kinds of AI explanation \textit{problems} faced. These problems include obtaining directly interpretable surrogate models and ``black-box explanation'', the latter of which is further divided into the problems of model explanation, outcome explanation, and model inspection. The paper further provides a list of algorithms to generate each type of explanation. Although more actionable, this paper is limited to the consideration of post-hoc explanation of black-box models and ignores other aspects such as understanding data~\cite{adadi2018peeking,hohman2019gamut}. 

%Fundamentally, it is necessary to 
We have argued in Sections~\ref{sec:intro} and \ref{sec:taxonomy} for the necessity of mapping explainability methods to explanation problems because AI is being used in diverse contexts by diverse groups of people, who may have different needs in demanding more transparent AI. Several works capture these diverse motivations for explainability. For example, Doshi-Velez and Kim \cite{rsi} enumerate the needs for explainability including to gain scientific understanding, ensure safety in complex tasks, guard against discrimination, and identify mismatched objectives. Adadi and Berrada~~\cite{adadi2018peeking} summarize the goals of AI explainability methods as justifying a model's decisions to users, improving user control of the AI by avoiding errors, enabling users to better improve the AI, and helping people discover new knowledge. %The AIX360 toolkit aims to provide a comprehensive and actionable taxonomy to help AI researchers and practitioners to select explainability algorithms that accommodate these different explainability needs.

In addition to survey papers proposing taxonomies, there are multiple explainability toolkits available through open source. Some of the more popular ones are shown in Table~\ref{tab:compare} along with the categories of methods they possess. The clearest difference between AIX360 and these other toolkits is that AIX360 offers a wider spectrum of explainability methods, notably data explanations, metrics, and persona-specific explanations, the last referring to TED's ability to provide explanations in the language of the persona. Alibi~\cite{alibi} primarily possesses local explainability methods such as Anchors \cite{anchors}, contrastive explanations \cite{CEM} and counterfactuals \cite{counterfactual}. Skater~\cite{skater} has directly interpretable models such as rule-based models and decision trees. These can also be used for global post-hoc interpretation. It also has LIME-like \cite{lime} methods for local post-hoc explainability. H2O~\cite{h2o} and InterpretML~\cite{interpret} both are similar to Skater in that they have methods that provide direct, local and global post-hoc explainability. The difference is in the particular methods they possess under these categories.
Both EthicalML-XAI~\cite{ethicalml} and DALEX~\cite{dalex} have different ways to visualize model behavior in terms of input variables, which could provide insight into global behavior. DALEX also has standard local explainability methods such as LIME and SHAP. %, and similar to EthicalML it also possesses ways to visualize different variables that 
tf-explain~\cite{tfexplain} and iNNvestigate~\cite{innvestigate}  primarily contain methods that provide local explanations for images. tf-explain also has ways to visualize activations of neurons of a neural network which could aid in understanding its global behavior.

Beyond differences in methods offered, AIX360 also distinguishes itself from other toolkits in terms of educational materials. In particular, AIX360 has materials aimed at less technical consumers, notably the web demo described in Section~\ref{sec:enhance:demo} which appears to be unique, as well as the guidance and glossary noted in Section~\ref{sec:taxonomy}. AIX360 also targets industry-specific users with the tutorials and web demo in Section~\ref{sec:enhance:demo}. Another distinctive feature of AIX360 is its common architecture and programming interface.

\section{Discussion}
\label{disc}

In summary, we have introduced the open-source AI Explainability 360 (AIX360) toolkit, which contains eight state-of-the-art explainability algorithms that can explain an AI model or a complex dataset in different ways to a diverse set of users. Some of the algorithms were enhanced from their published versions for AIX360 to make them more usable and their outputs easier to consume. The toolkit also contains two explainability metrics, %all implemented with an easily extensible API. These aspects distinguish it from other existing explainability toolkits making it more comprehensive than them. The toolkit also has 
a credit approval demonstration, five elaborate tutorials covering different real-world use cases, and 13 Jupyter notebooks, making it accessible to practitioners and non-experts. These resources, together with the breadth of the toolkit and its common extensible programming interface, help distinguish AIX360 from existing explainability toolkits.

We have also provided a simple yet comprehensive taxonomy that structures the explainability space and serves multiple purposes, including guiding practitioners, revealing algorithmic gaps to researchers, and informing design choices related to explainability software (e.g.,~class hierarchy) for data scientists and developers. We thus believe that our toolkit and taxonomy provide technical, educational, and operational benefits to the community over and above what currently exists. As AI surges forward with trust being one of the critical bottlenecks in its widespread adoption \cite{trustbottleneck}, we hope that the community will leverage and significantly enhance our contributions.

%The toolkit is easily extensible as described before, soliciting many more contributions related to explainable AI. 
The AIX360 toolkit solicits contributions related to explainable AI to enable it to grow.
This includes the categories discussed in Section~\ref{sec:implement}, where extensions can be easily handled. %but is not limited to explainability algorithms for local and global explainability and explainability metrics along with unsupervised algorithms that help in understanding complex datasets. Saying this though, 
Of particular interest are contributions to categories that are present in the taxonomy of Figure~\ref{fig:algo-guidance} but not in the toolkit. These are indicated in Figure \ref{fig:algo-guidance} by ``?'' in three cases: interactive, static $\rightarrow$ data $\rightarrow$ distributions, and static $\rightarrow$ model $\rightarrow$ global $\rightarrow$ post-hoc $\rightarrow$ visualize. There is a significant literature in the `visualize' category, which includes visualizing intermediate nodes and/or layers in a deep neural network \cite{nguyen2016multifaceted,deepling} as well as plotting the effects of input features on the output \cite{molnarbook} to assess their importance. Such contributions would be highly relevant for the toolkit. There is comparatively much less work in the other two categories, notwithstanding calls for interactive explanation~\cite{miller2018explanation,hohman2019gamut,weld2018intelligible}, and we hope that the taxonomy inspires %this highlights the need inspiring 
more research in those directions. %Interactive is especially significant, as humans typically ask multiple related questions before they are satisfied with an explanation. 
Another area where contributions would be highly welcome are for modalities that are not covered under the categories that do have algorithms (e.g.,~contrastive for text). Contributions in categories that are missing from the taxonomy in Figure~\ref{fig:algo-guidance} are of course valuable as well.

Lastly, a large variety of deep learning frameworks are available today to train and store models (e.g., TensorFlow, Keras, PyTorch, MXNet). Therefore explainability toolkits should allow users to explain models that have been built using different frameworks while avoiding the need to implement explainability algorithms multiple times for each framework. To ease the task of algorithm developers and make explainability algorithms framework-independent, future work should implement framework-specific model classes that expose a common API needed by explainability algorithm developers. Preliminary work in this direction was initiated in AIX360 by defining a class for Keras-based classifiers that is utilized by CEMExplainer and CEMMAFImageExplainer. 

\section*{Acknowledgment}

We would like to thank MaryJo Fitzgerald for assistance with the web demo, Karthikeyan Natesan Ramamurthy for assistance with the Medical Expenditure Panel Survey data, and Joanna Brewer for copy editing.

\bibliographystyle{abbrv}
\bibliography{aix} 

\ifappendix
\clearpage
\appendix
\section{Accuracy-Complexity Trade-offs for BRCG-light}
\label{sec:BRCGexpt}

Figures~\ref{fig:paretoAll1}--\ref{fig:paretoAll3} show trade-offs between classification accuracy and DNF rule complexity for BRCG-light (described in Section~\ref{sec:enhance:BRCG} and labelled BRCG-l in the figure legends) in comparison to the integer programming version of BRCG (BRCG-IP), Bayesian Rule Sets (BRS) \cite{wang2017bayesian}, and the alternating minimization (AM) and block coordinate descent (BCD) algorithms from \cite{su2016learning}. For BRCG-light, three settings of the beam search width are evaluated: $B = 1, 3, 5$. The $15$ datasets considered here are the same as in \cite{BDR}. Complexity is measured as the number of clauses in the DNF rule (equivalently the number of rules in the rule set) plus the total number of conditions in all clauses/rules, also following \cite{BDR}. Overall, the comparisons show that BRCG-light provides trade-offs that are worse on average than BRCG-IP, as may be expected given the dramatic reduction in computation, but better than the other methods. They also show that $B = 5$ gives better results than smaller widths on most datasets, particularly the larger ones.

\begin{figure*}[ht]
  \centering
  \begin{subfigure}[b]{0.45\textwidth}
  \includegraphics[width=\textwidth]{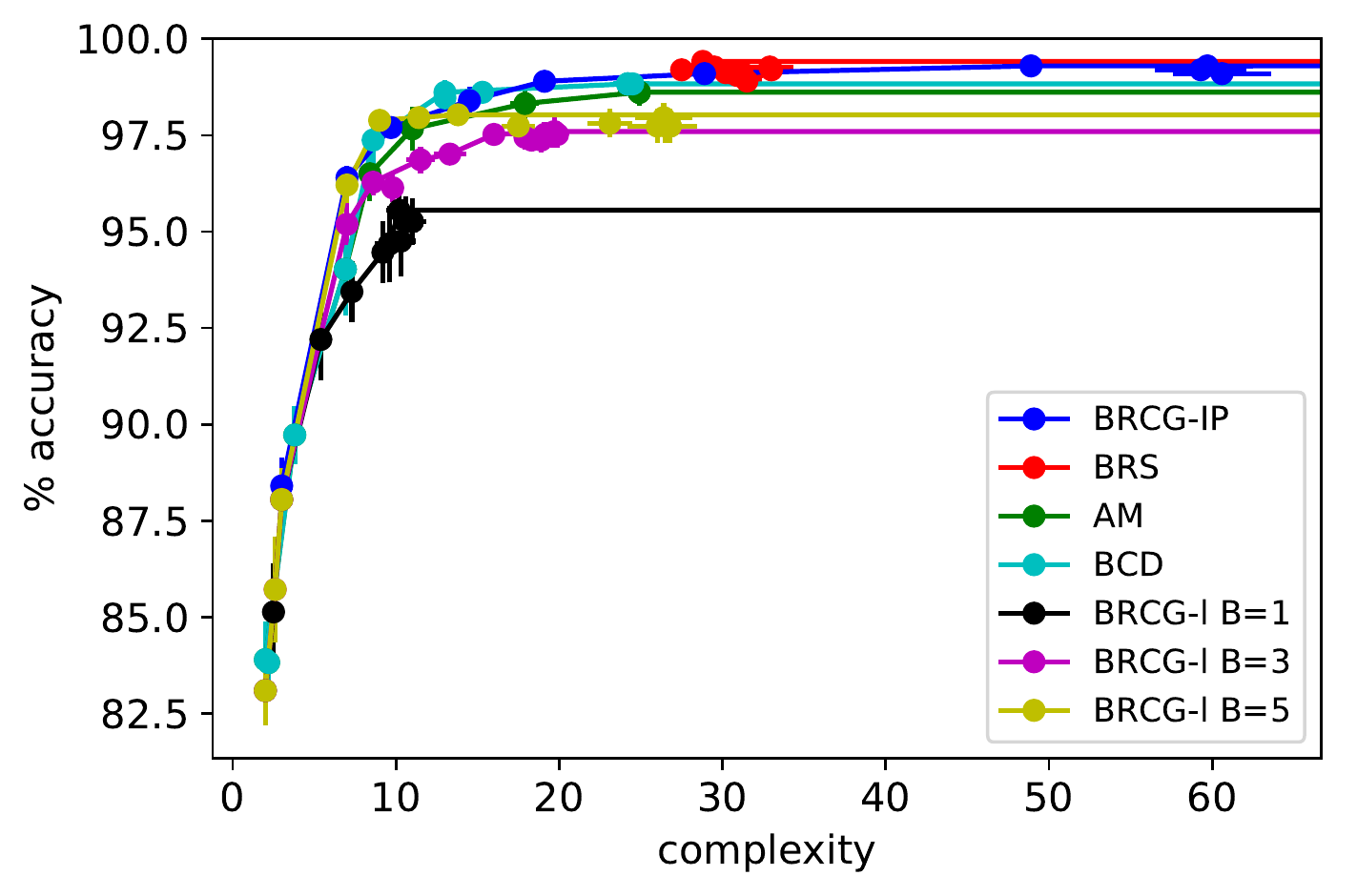}
  \caption{banknote}
  \end{subfigure}
  \begin{subfigure}[b]{0.45\textwidth}
  \includegraphics[width=\textwidth]{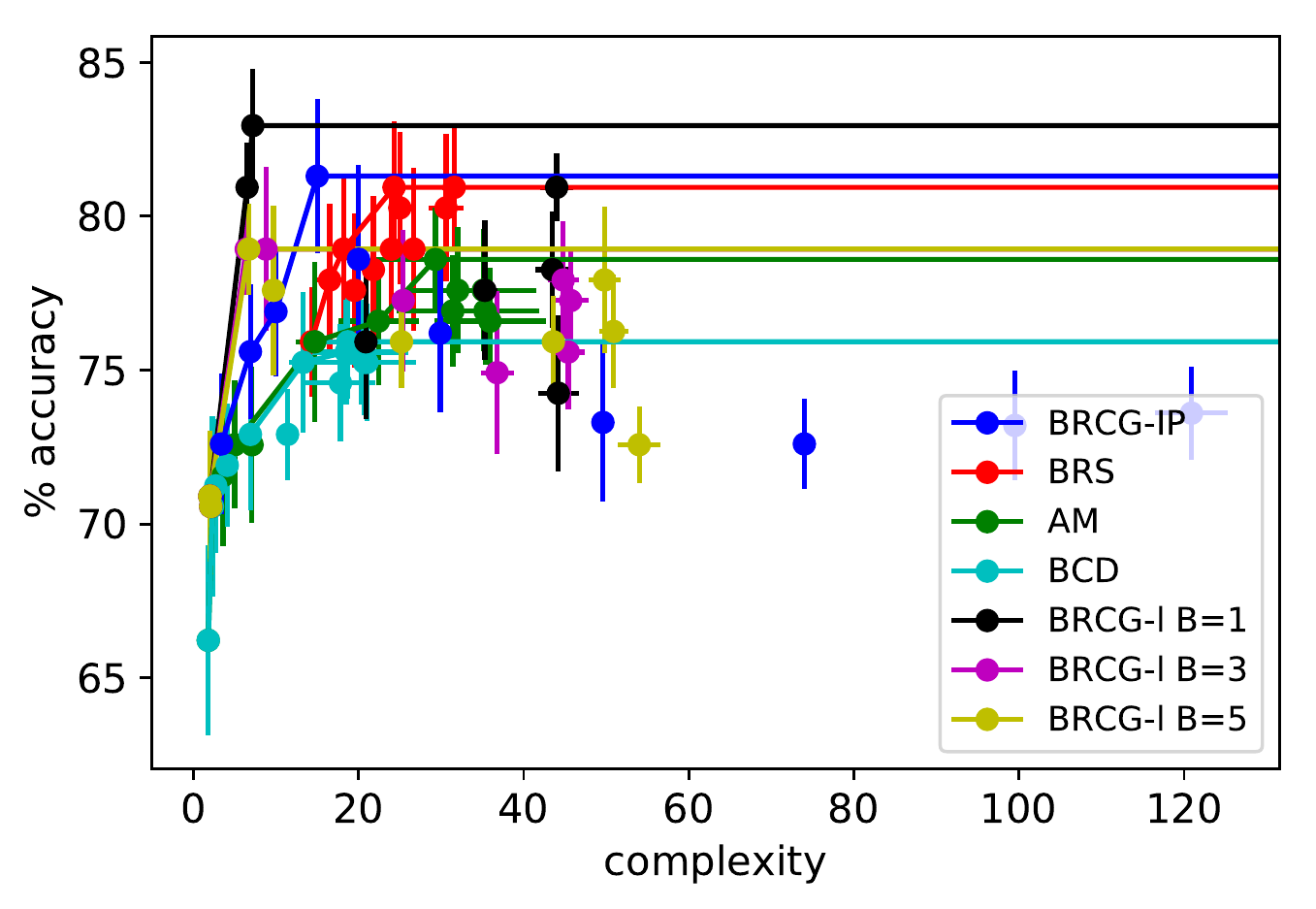}
  \caption{heart}
  \end{subfigure}
  \begin{subfigure}[b]{0.45\textwidth}
  \includegraphics[width=\textwidth]{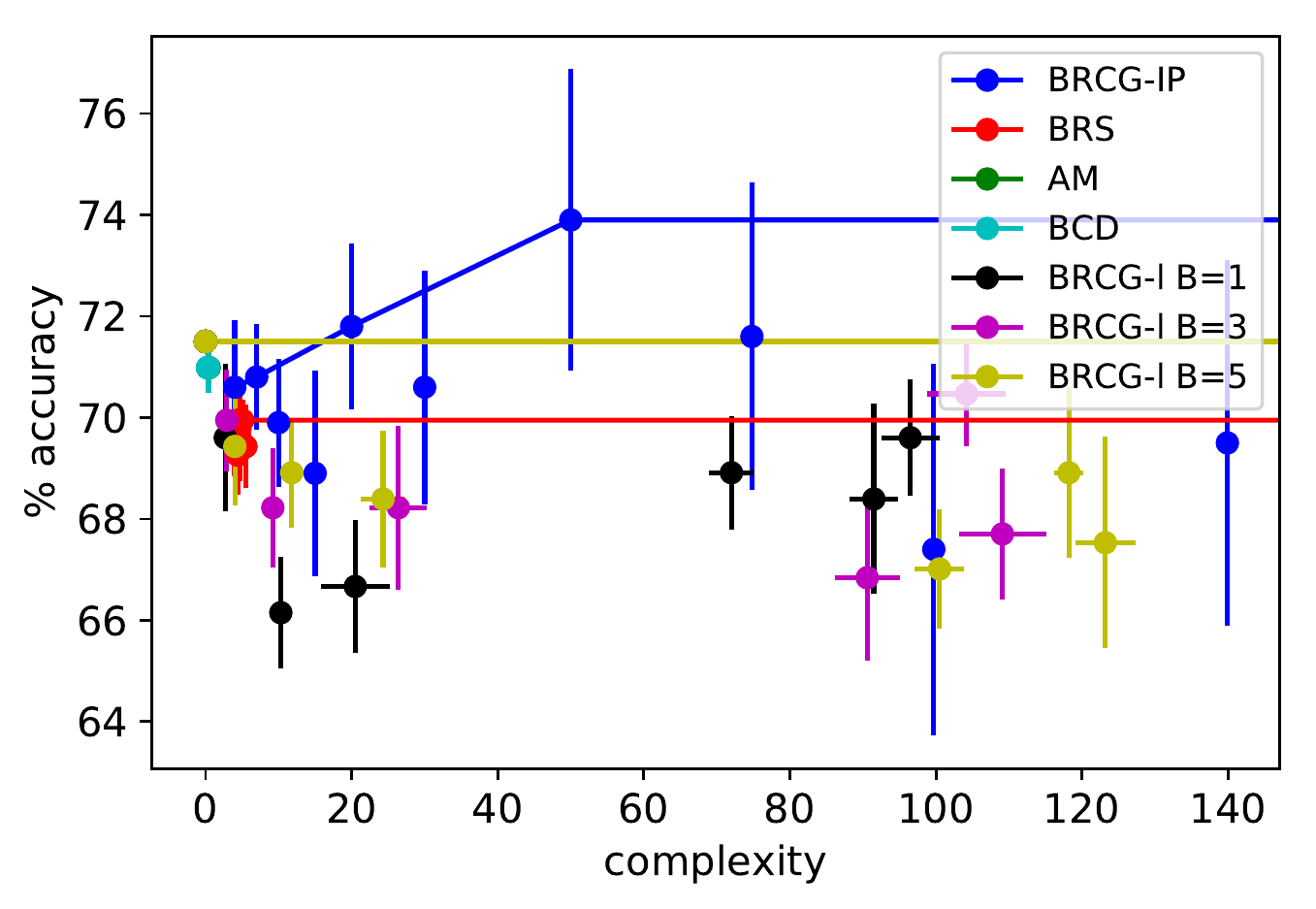}
  \caption{ILPD}
  \end{subfigure}
  \begin{subfigure}[b]{0.45\textwidth}
  \includegraphics[width=\textwidth]{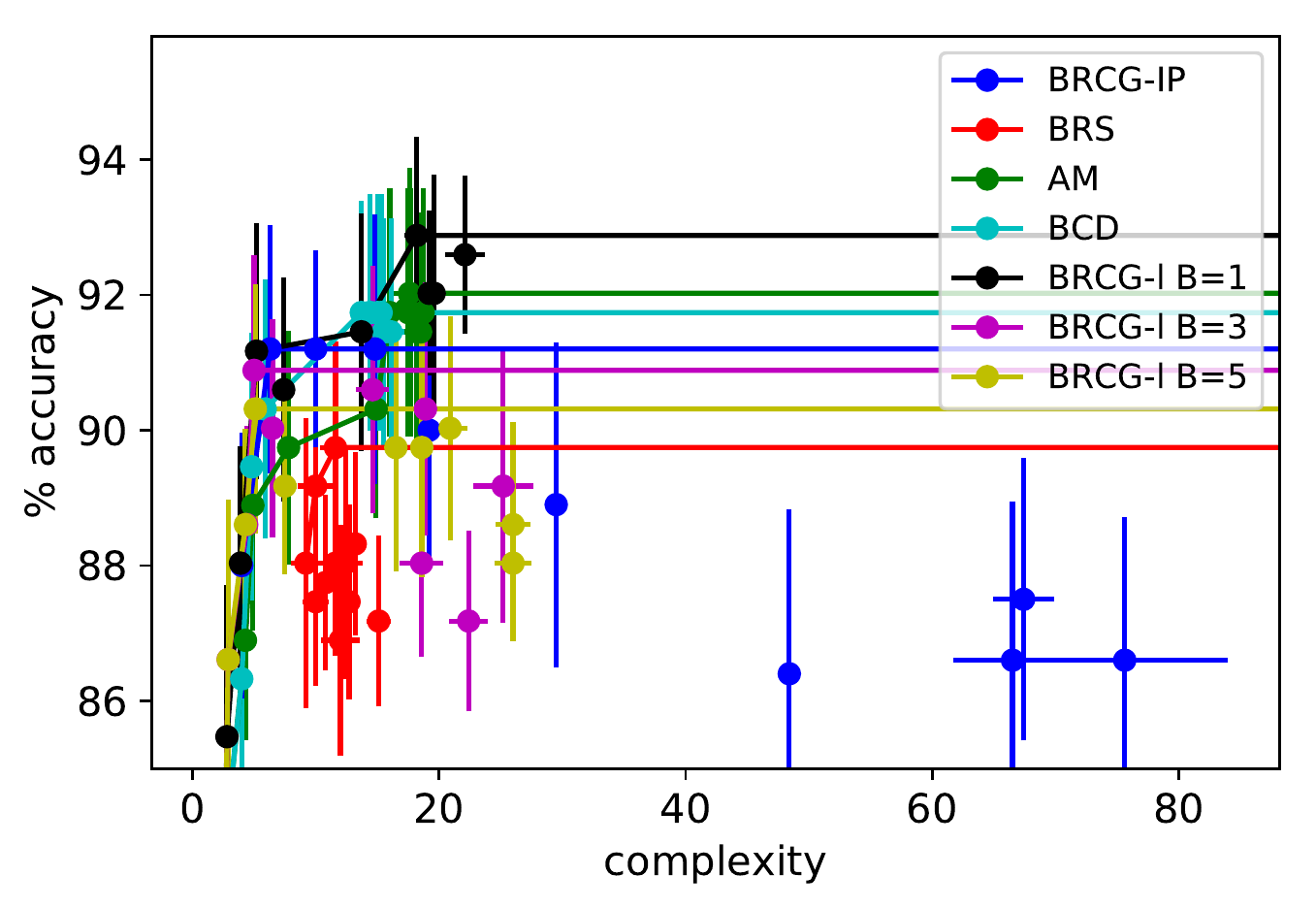}
  \caption{ionosphere}
  \end{subfigure}
  \begin{subfigure}[b]{0.45\textwidth}
  \includegraphics[width=\textwidth]{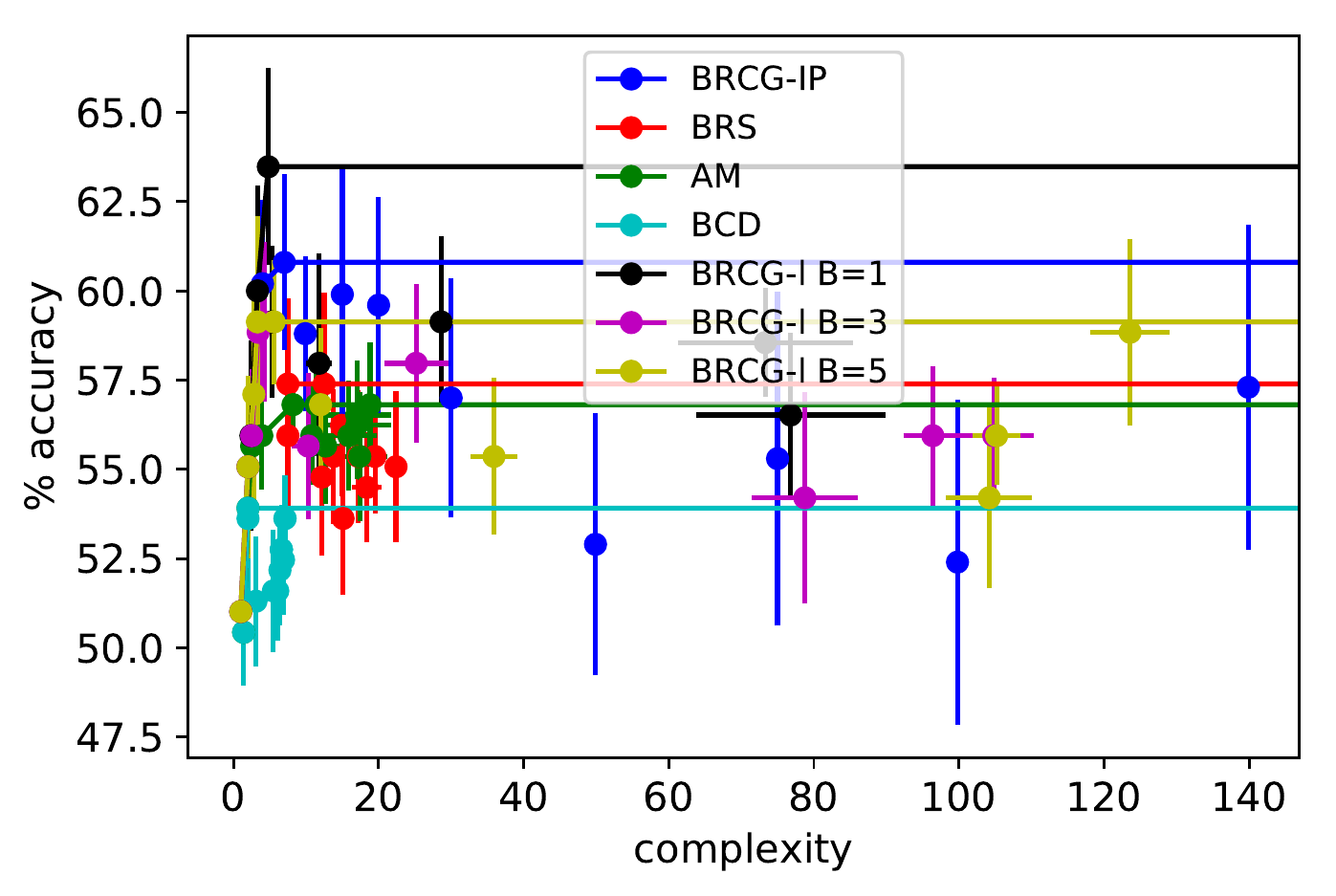}
  \caption{liver}
  \end{subfigure}
  \begin{subfigure}[b]{0.45\textwidth}
  \includegraphics[width=\textwidth]{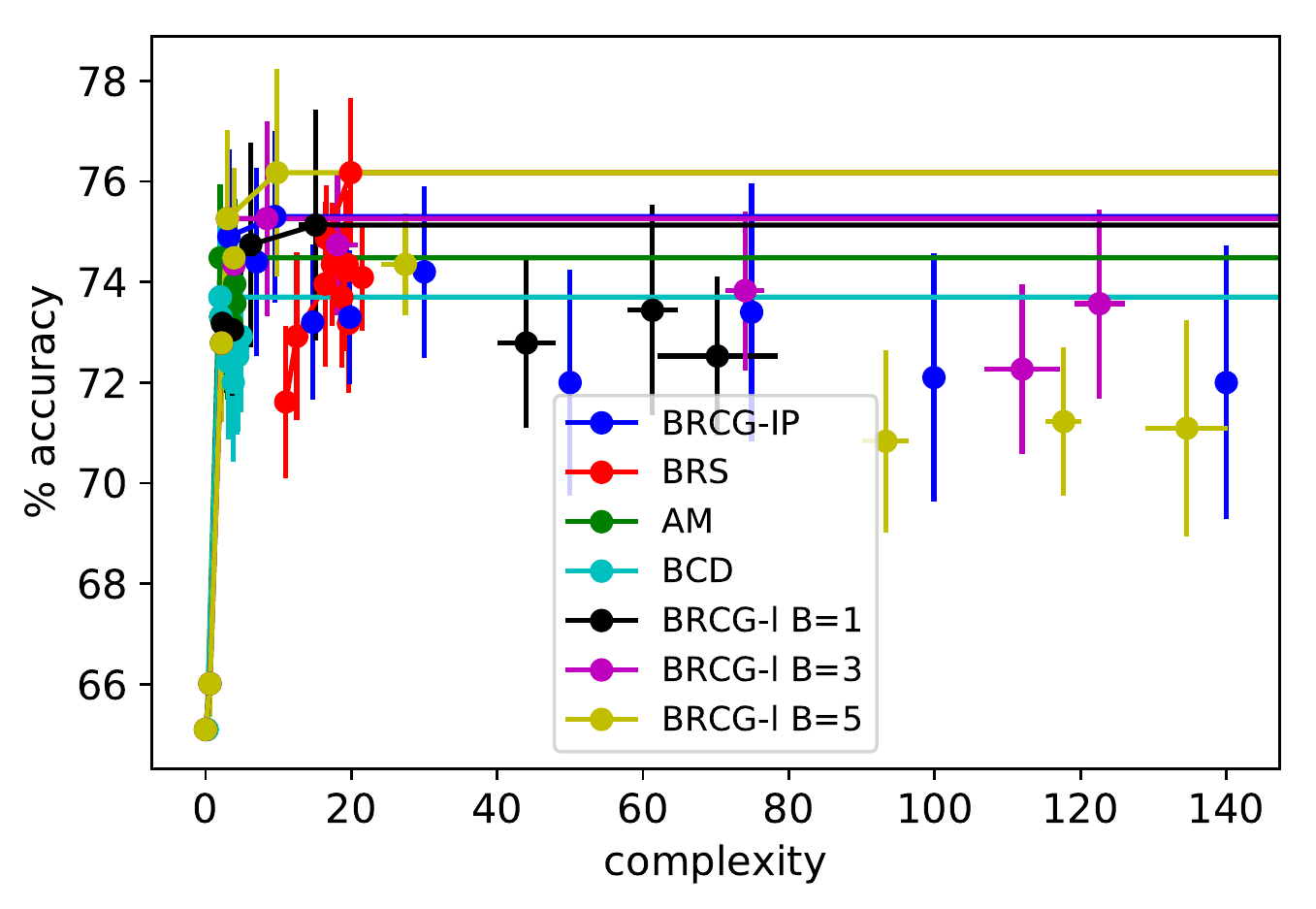}
  \caption{pima}
  \end{subfigure}
  \caption{Rule complexity-test accuracy trade-offs. Pareto efficient points are connected by line segments. Horizontal and vertical bars represent standard errors in the means.}
  \label{fig:paretoAll1}
\end{figure*}

\begin{figure*}[ht]
  \centering
  \begin{subfigure}[b]{0.45\textwidth}
  \includegraphics[width=\textwidth]{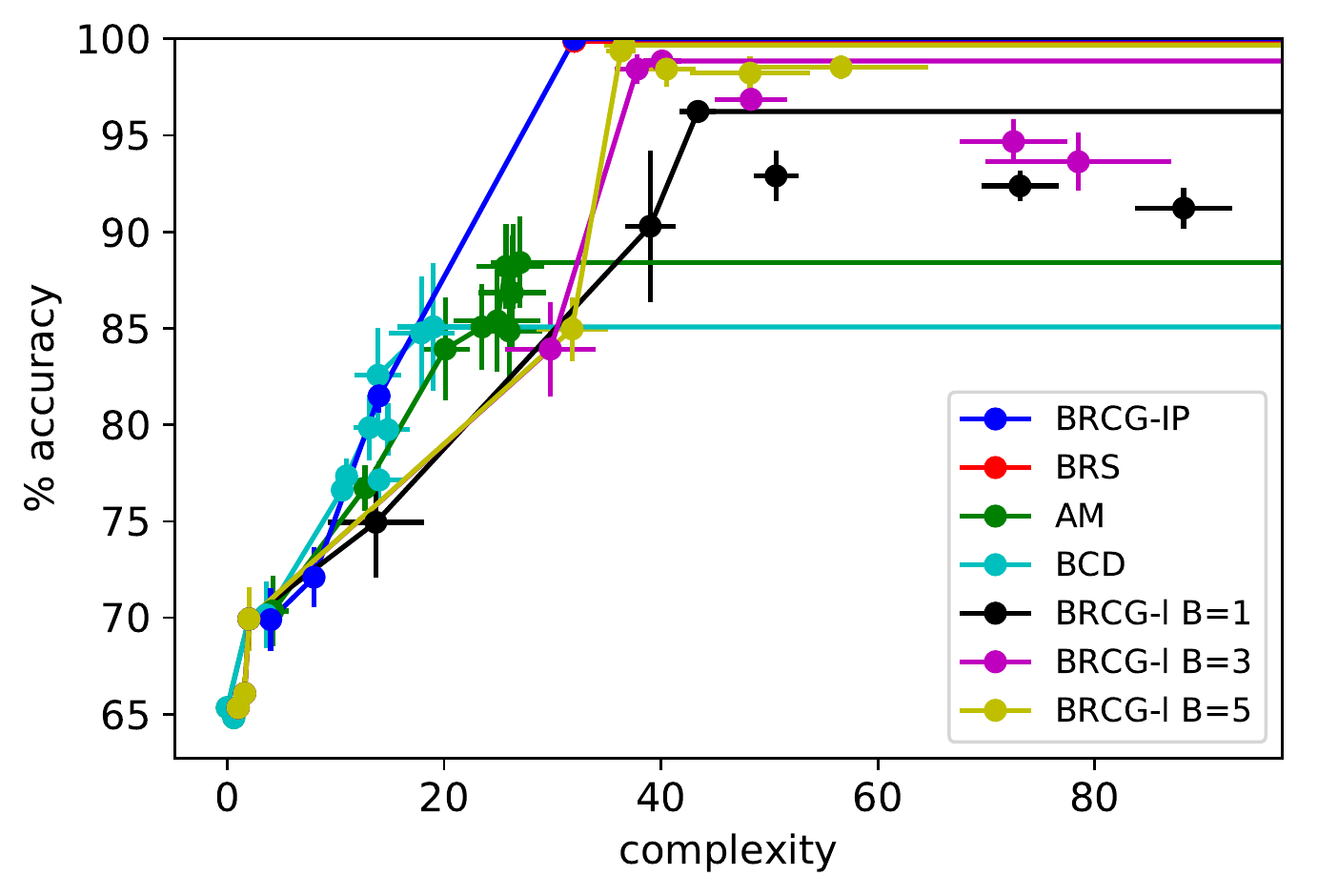}
  \caption{tic-tac-toe}
  \end{subfigure}
  \begin{subfigure}[b]{0.45\textwidth}
  \includegraphics[width=\textwidth]{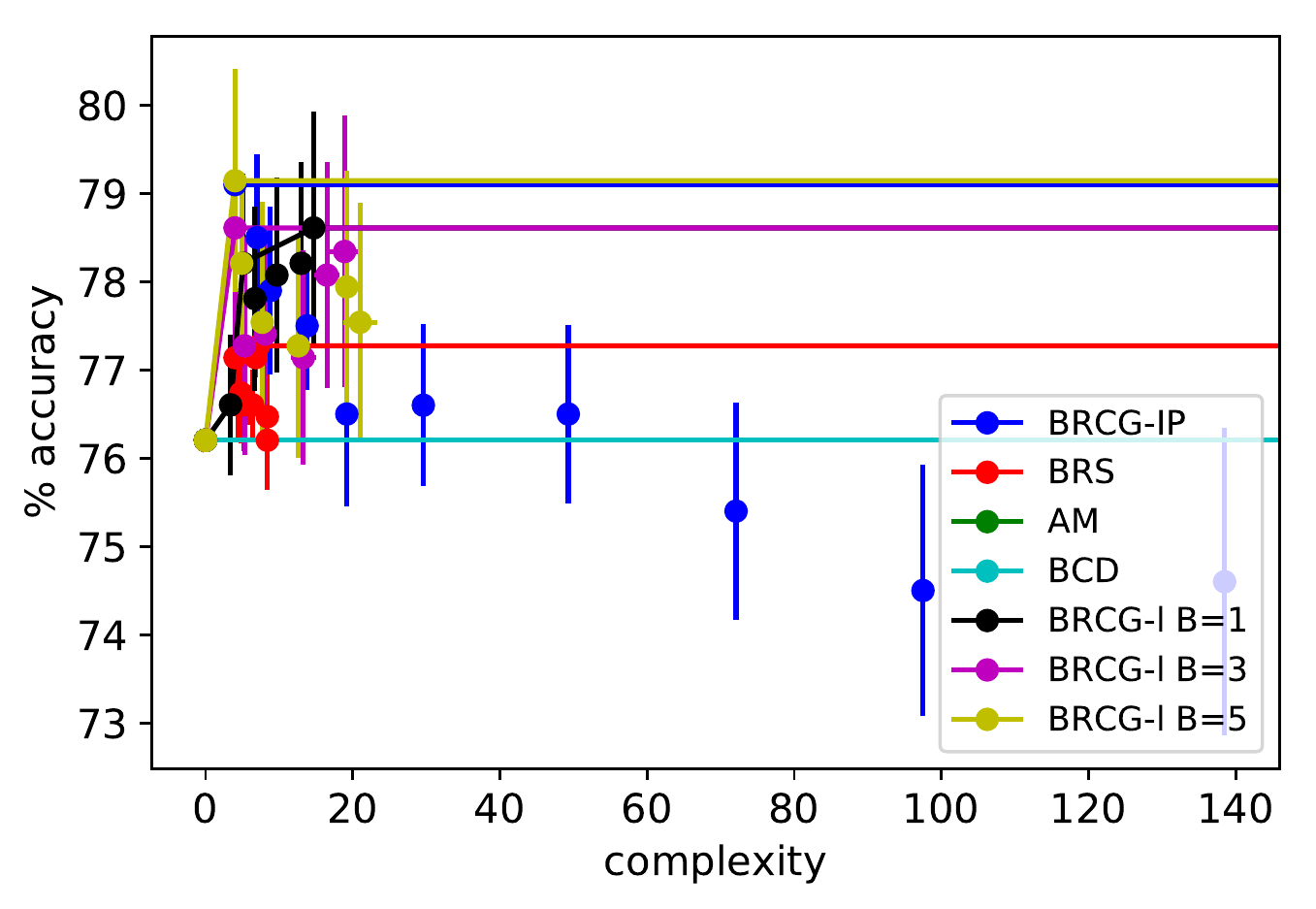}
  \caption{transfusion}
  \end{subfigure}
  \begin{subfigure}[b]{0.45\textwidth}
  \includegraphics[width=\textwidth]{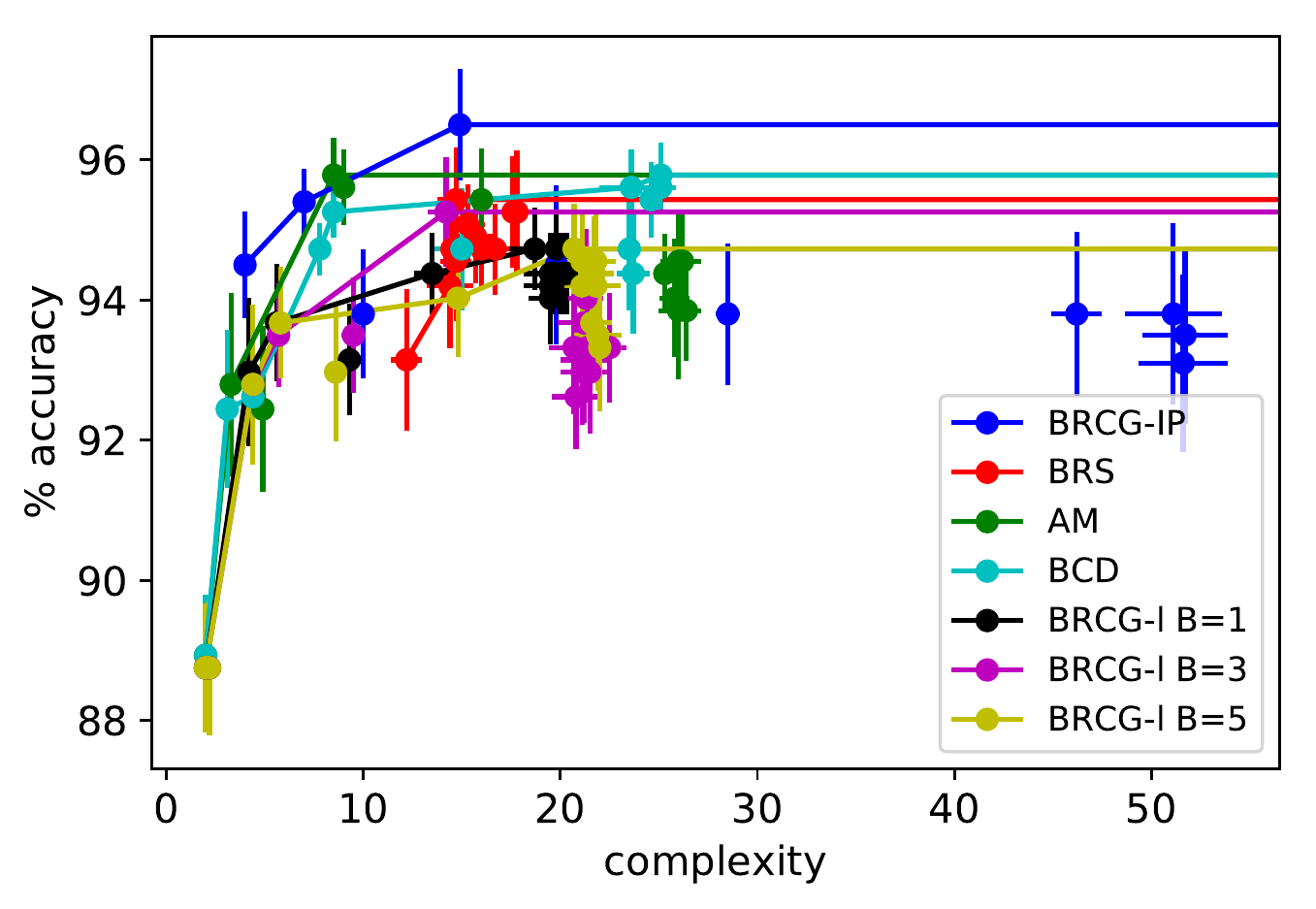}
  \caption{WDBC}
  \end{subfigure}
  \begin{subfigure}[b]{0.45\textwidth}
  \includegraphics[width=\textwidth]{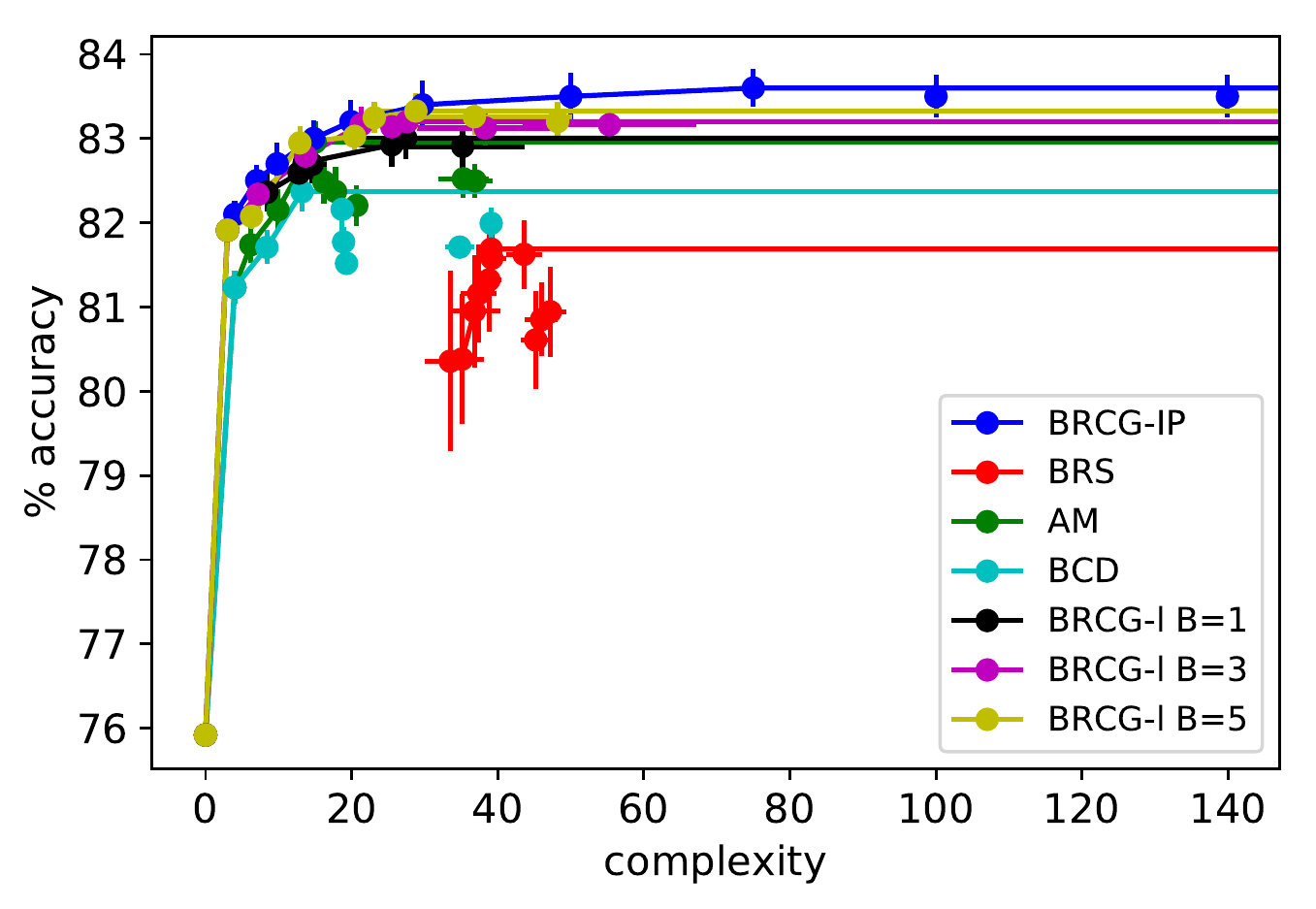}
  \caption{adult}
  \end{subfigure}
  \begin{subfigure}[b]{0.45\textwidth}
  \includegraphics[width=\textwidth]{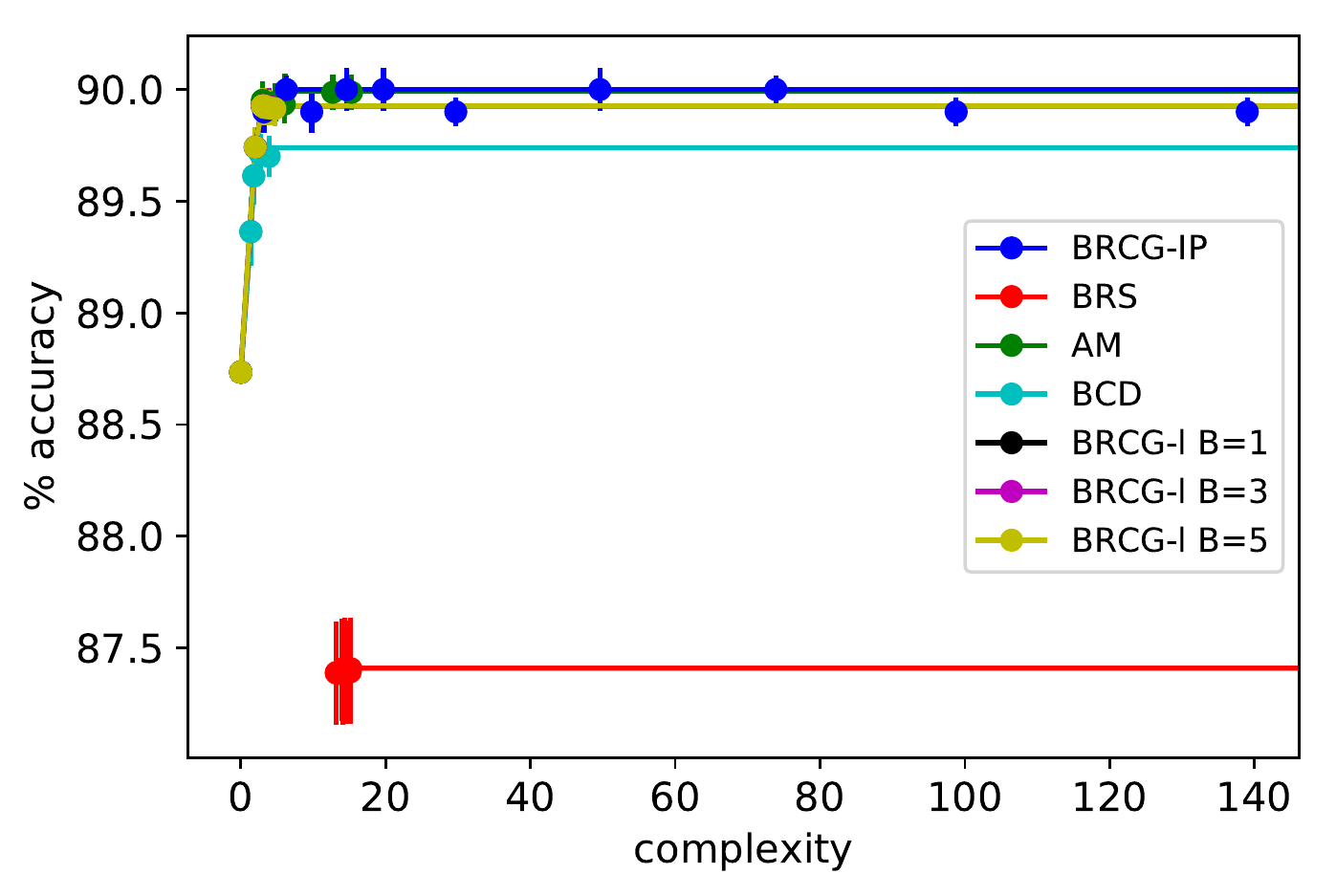}
  \caption{bank-marketing}
  \end{subfigure}
  \begin{subfigure}[b]{0.45\textwidth}
  \includegraphics[width=\textwidth]{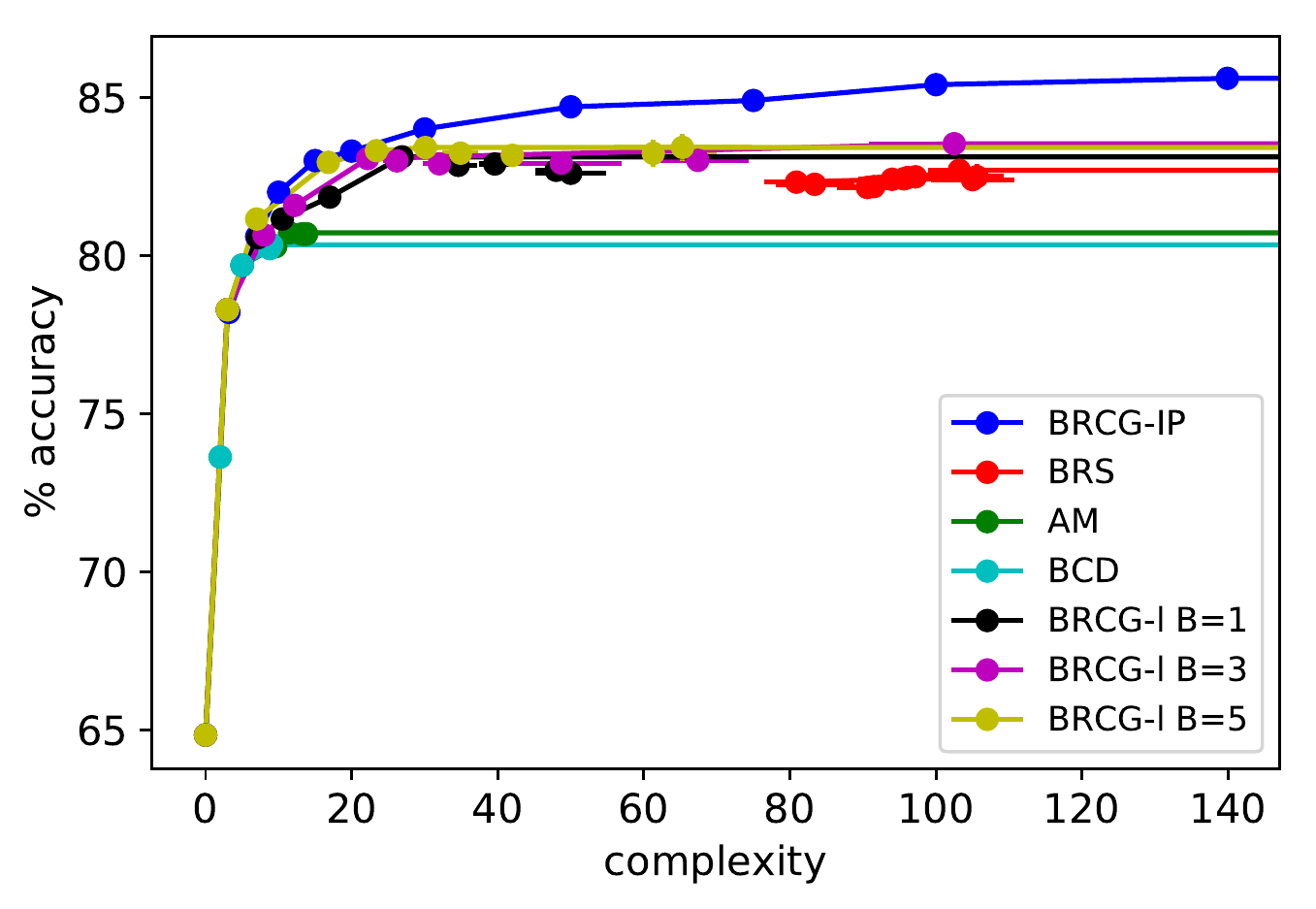}
  \caption{magic}
  \end{subfigure}
  \caption{Rule complexity-test accuracy trade-offs. Pareto efficient points are connected by line segments. Horizontal and vertical bars represent standard errors in the means.}
  \label{fig:paretoAll2}
\end{figure*}

\begin{figure*}[ht]
  \centering
  \begin{subfigure}[b]{0.45\textwidth}
  \includegraphics[width=\textwidth]{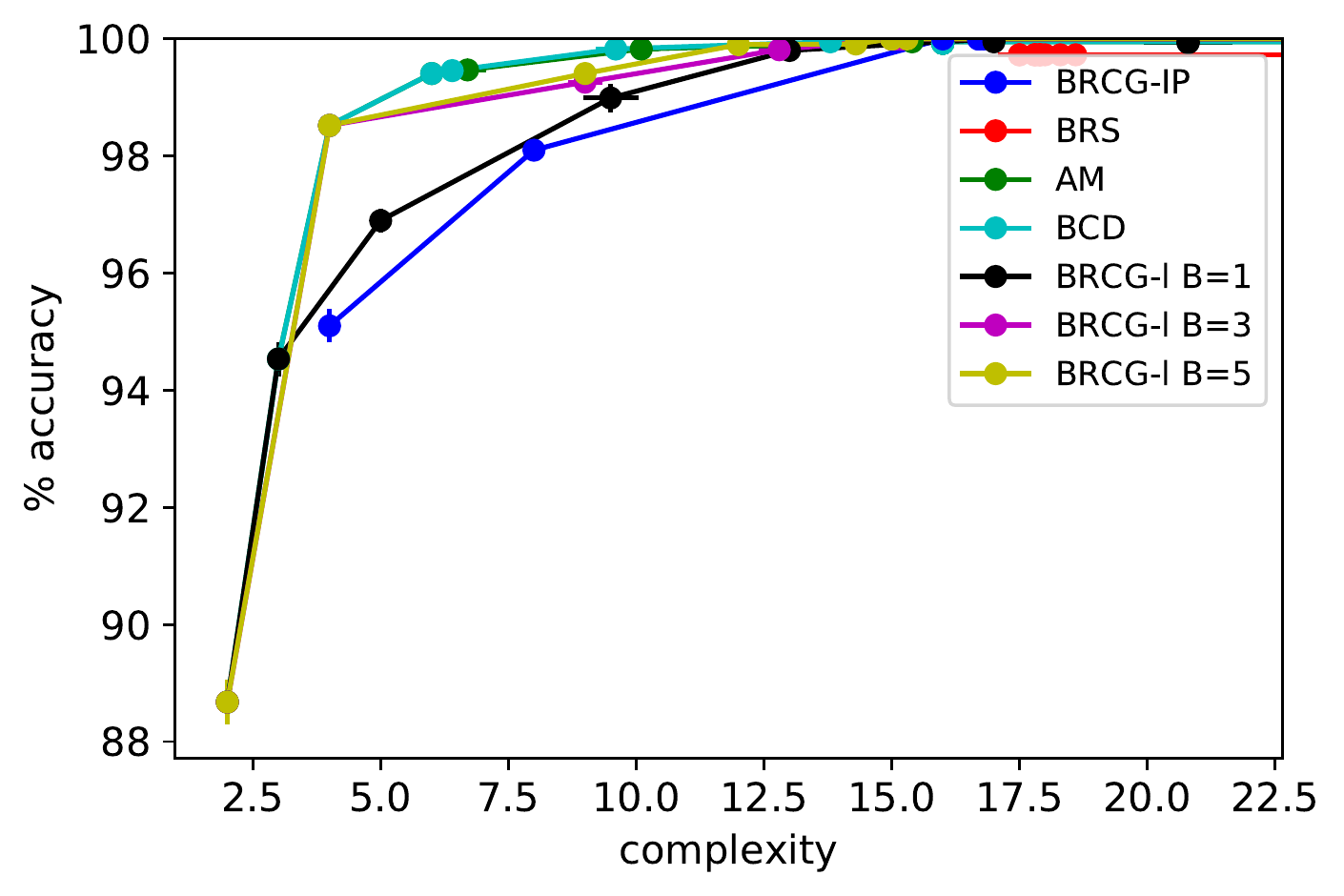}
  \caption{mushroom}
  \end{subfigure}
  \begin{subfigure}[b]{0.45\textwidth}
  \includegraphics[width=\textwidth]{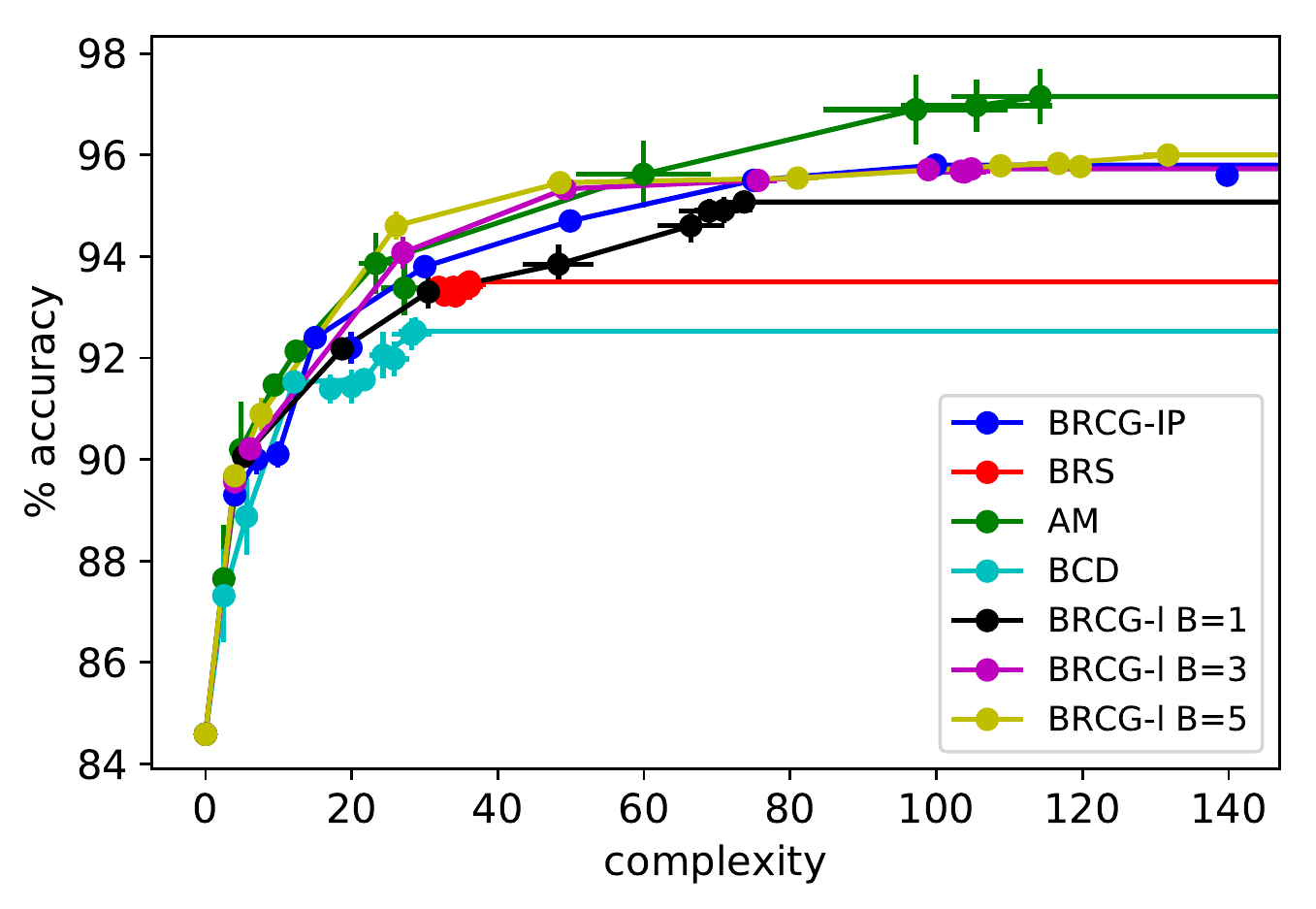}
  \caption{musk}
  \end{subfigure}
  \begin{subfigure}[b]{0.45\textwidth}
  \includegraphics[width=\textwidth]{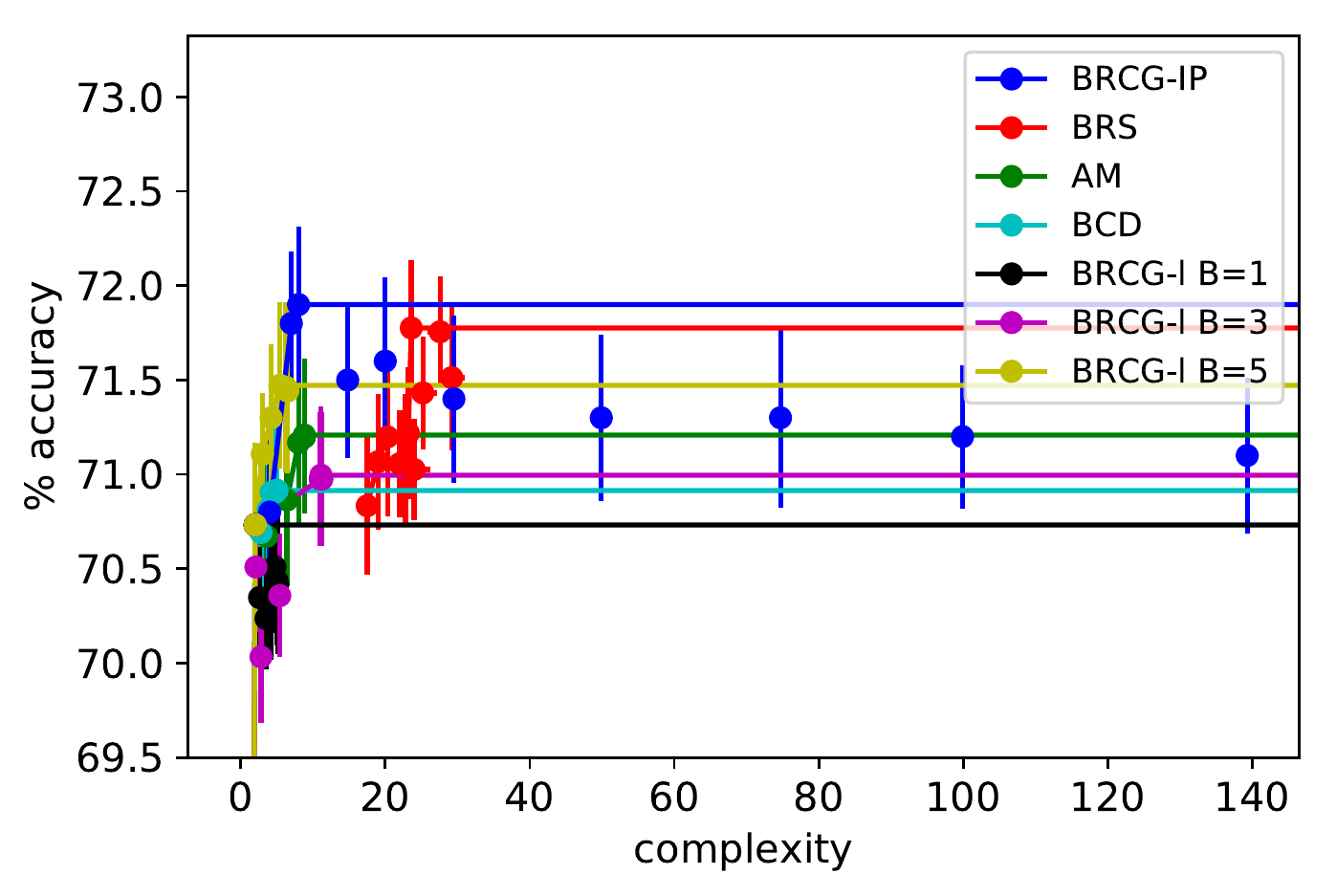}
  \caption{FICO}
  \end{subfigure}
  \caption{Rule complexity-test accuracy trade-offs. Pareto efficient points are connected by line segments. Horizontal and vertical bars represent standard errors in the means.}
  \label{fig:paretoAll3}
\end{figure*}

\clearpage

\begin{figure*}[ht]
  \centering
  \begin{subfigure}[b]{0.8\textwidth}
  \includegraphics[width=\textwidth]{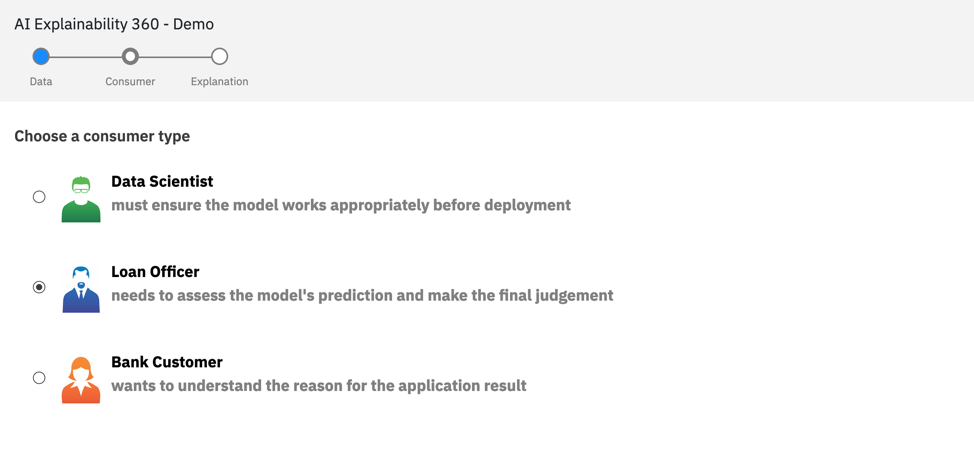}
  \caption{three types of explanation consumers in the FICO use case}
  \end{subfigure}
  \newline
  \newline
  \newline
  \newline
  \begin{subfigure}[b]{0.8\textwidth}
  \includegraphics[width=\textwidth]{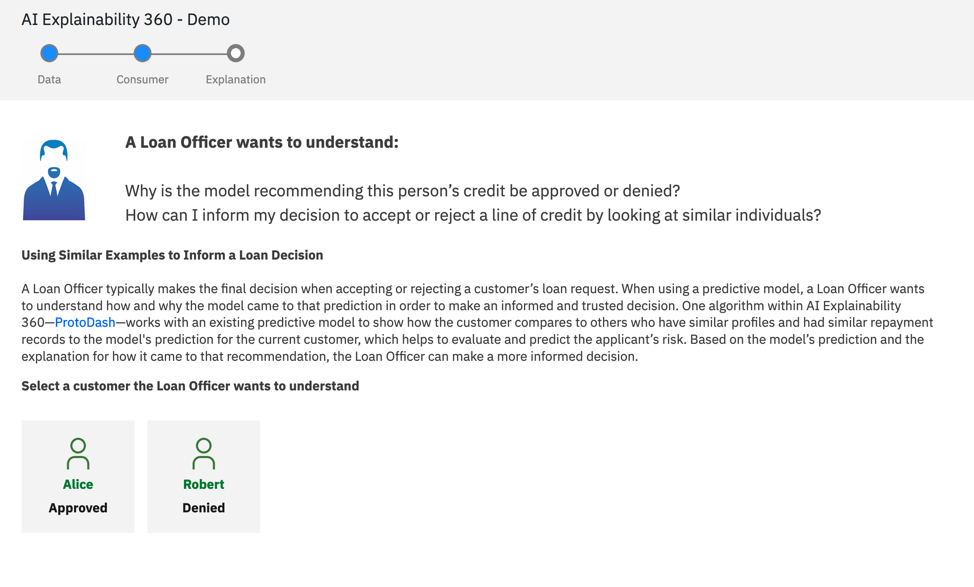}
  \caption{the loan officer can choose either customer to see an explanation}
  \end{subfigure}

 \end{figure*}
\clearpage

\begin{figure*}[ht]\ContinuedFloat
  \centering
    \begin{subfigure}[b]{0.7\textwidth}
  \includegraphics[width=\textwidth]{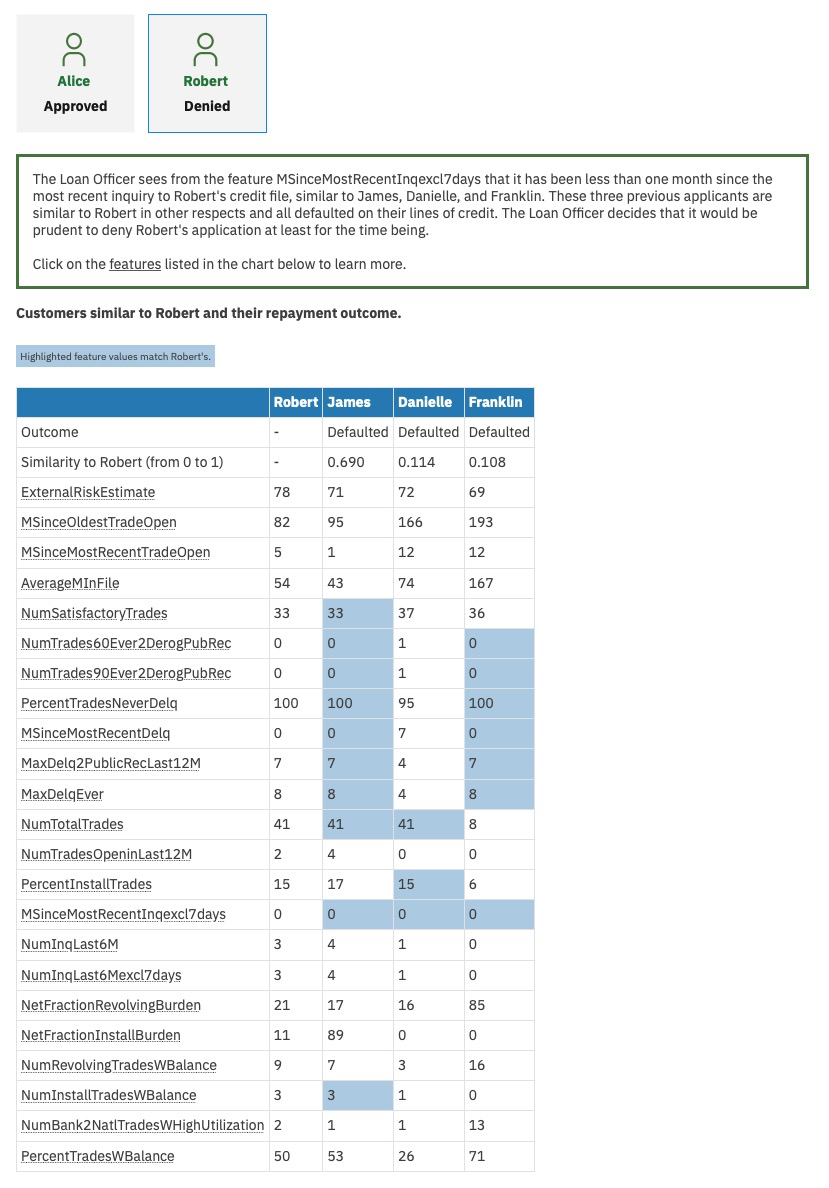}
  \caption{explanation for Robert generated by ProtoDash}
  \end{subfigure}
  \newline
  \newline
  \caption{Web demo based on FICO Explainable Machine Learning Challenge. Three types of explanation consumers are involved in this demo and their needs are best served by different types of explainability methods.}
  \label{fig:demo}
\end{figure*}
\clearpage

%\section{Web Demo Screenshots}
%\label{sec:demoScreenshots}

\fi

\end{document}